\documentclass{article}

\PassOptionsToPackage{numbers,sort}{natbib}

\usepackage[main, final]{neurips_2026}

\usepackage[utf8]{inputenc} 
\usepackage[T1]{fontenc}    
\usepackage{hyperref}       
\hypersetup{
  colorlinks=true,
  linkcolor=[RGB]{51,94,150},
  citecolor=[RGB]{51,94,150},
  urlcolor=[RGB]{51,94,150},
}

\usepackage{graphicx}
\usepackage{tabularx}
\usepackage{url}            
\usepackage{booktabs}       
\usepackage{amsmath}
\usepackage{amsfonts}       
\usepackage{amssymb}
\usepackage{nicefrac}       
\usepackage{microtype}      
\usepackage{needspace}
\usepackage[table,dvipsnames]{xcolor}  
\definecolor{TableHeaderLavender}{RGB}{235,232,243}
\definecolor{TableStripeSoft}{RGB}{241,241,241}
\definecolor{TableStripeGreen}{rgb}{.961,.985,.970}
\definecolor{TableStripePurple}{RGB}{239,233,250}
\newcommand{\gaingreen}[1]{$_{\color{green!45!black}\uparrow #1}$}
\newcommand{\gainblue}[1]{$_{\color{blue!65!black}\uparrow #1}$}
\newcommand{\gainblack}[1]{\raisebox{-0.6ex}{\textcolor{black}{\scriptsize\itshape #1}}}
\newcommand{\qcot}{Qwen3.5-9B\gainblack{CoT}\ }
\newcommand{\qsft}{Qwen3.5-9B\gainblack{SFT+GRPO}\ }
\newcommand{\latentcachemodel}{\textcolor[RGB]{61,24,196}{\textbf{\textsc{Latent-VC-9B}}}}
\usepackage{pifont}
\usepackage[ruled,vlined,linesnumbered]{algorithm2e}
\DontPrintSemicolon
\SetKwComment{SideComment}{\#\ }{}
\newcommand{\algblue}[1]{\SideComment*[r]{\textcolor{blue!65!black}{#1}}}
\newcommand{\alggreen}[1]{\SideComment*[r]{\textcolor{green!45!black}{#1}}}
\newcommand{\algred}[1]{\SideComment*[r]{\textcolor{red!65!black}{#1}}}

\usepackage{etoolbox}
\makeatletter
\patchcmd{\@algocf@start}{-1.5em}{-1pt}{}{}
\makeatother

\title{Latent Visual Cache for Video Reasoning}

\author{
  \textbf{Yongheng Zhang}$^{1,2*}$ \quad
  \textbf{Zhipeng Xu}$^{2*}$ \quad
  \textbf{Hao Wu}$^{2*}$ \quad \\
  \textbf{Yinghui Li}$^{1,2\dagger}$ \quad
  \textbf{Di Yin}$^{1}$ \quad
  \textbf{Xing Sun}$^{1}$ \quad
  \vspace{2mm}
  \textbf{Philip S. Yu}$^{3}$ \\
  $^1$Tencent Youtu Lab \\
  $^2$Tsinghua University \quad
  $^3$University of Illinois at Chicago
  \vspace{-5mm}
}

\begin{document}

\maketitle
\begingroup
\renewcommand{\thefootnote}{}
\footnotetext{$^*$ Equal Contribution. \quad $^\dagger$ Corresponding Author.}
\endgroup

\begin{abstract}
Video reasoning requires Large Multimodal Models (LMMs) to remain grounded in dense evidence, yet existing systems largely adopt ``read-once, generate-many'' paradigm, in which visual grounding weakens during generation. This phenomenon has been widely observed and is known as \textit{Visual Anchoring Decay}. To fill this gap, we introduce \textbf{Latent Video Cache} (\textsc{Latent-VC}), a recurrent latent visual cache inserted into the decoder to preserve compact visual memories throughout reasoning. The cache is trained with supervised contrastive cache alignment and vision-grounded GRPO with a latent grounding reward, while maintaining strict train-inference alignment through native decoder hidden states. Built on Qwen3.5-9B, \textsc{Latent-VC} consistently outperforms strong CoT and SFT+GRPO baselines across six video benchmarks, with especially clear gains on grounding-intensive and long-video tasks. In addition, it also achieves higher accuracy with substantially shorter responses, suggesting that latent visual caching improves video reasoning by preserving visual evidence rather than relying on longer textual chains.
\end{abstract}

\section{Introduction}

Video reasoning represents a critical frontier in multimodal reasoning, serving as the bedrock for interpreting the dynamic physical world and enabling high-stakes applications like autonomous driving and robotics~\cite{reed2022generalist,zhou2024vision,jiang2023vad,brooks2024video,wan2025wan,li2022past,liu2026let,lu2025youtu}. While its staggering spatiotemporal density poses formidable challenges, the emergence of Large Multimodal Models (LMMs) has recently catalyzed a paradigm shift~\cite{team2025kimi, chen2024internvl,singh2025openai,comanici2025gemini,zhang2026chatbotdigitalcolleagueparadigm,li2026cognitive}. Leveraging massive pre-trained backbones, LMMs successfully bridge the semantic gap between static perception and dynamic video reasoning~\cite{maaz2024video,fu2025video,zhang2025videollama,wang2025internvl3}.

\begin{figure}[t]
	\centering
	\includegraphics[width=\textwidth]{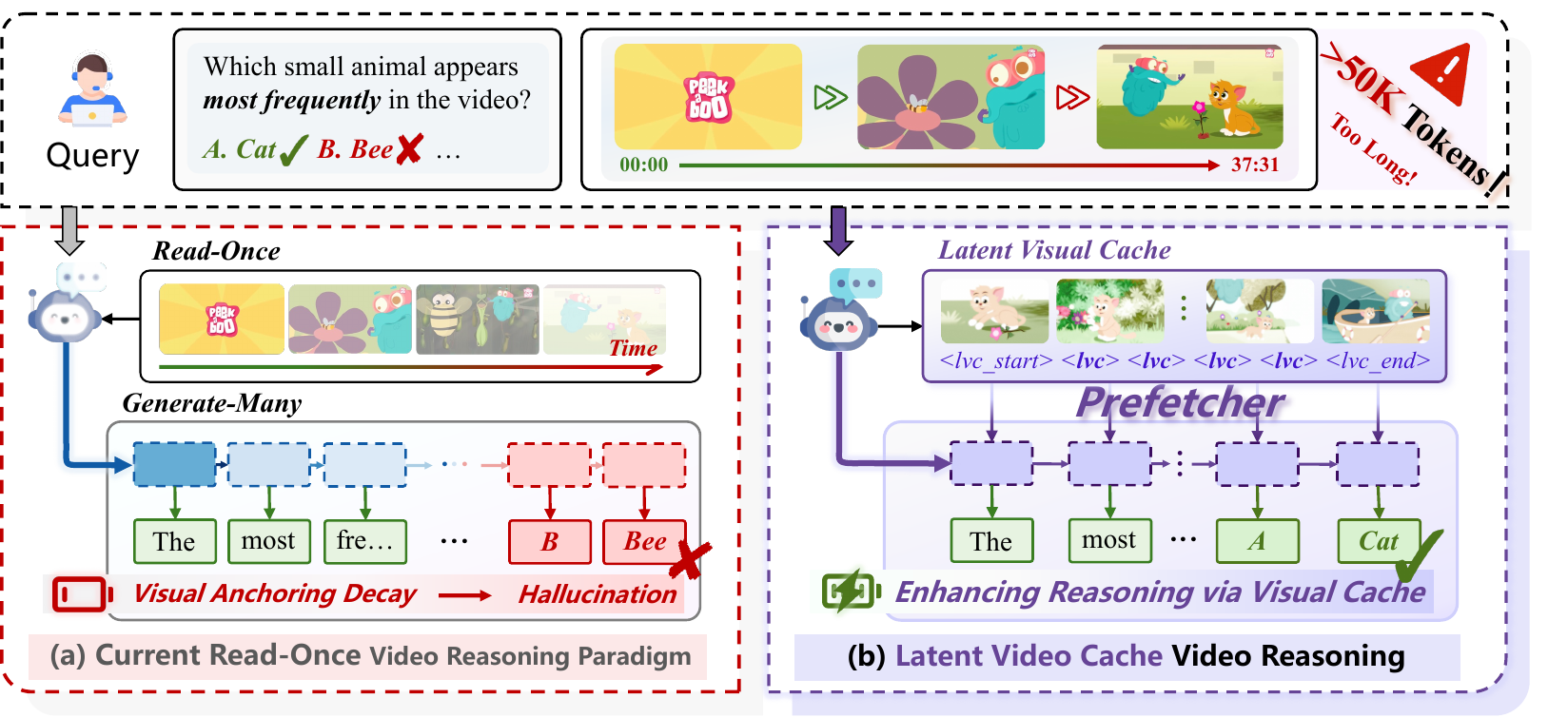}
	\caption{Comparison of long-video reasoning paradigms. (a) In the read-once paradigm, visual evidence is forgotten as the reasoning chain grows, leading to hallucinations. (b) In \textsc{Latent-VC}, a latent visual cache is constructed before generation to preserve grounding throughout reasoning.
 \vspace{-5mm}}
	\label{fig:intro}
\end{figure}

Despite this progress, LMMs are hindered by the ``\textit{\textbf{Read-Once, Generate-Many}}'' paradigm~\cite{zhang-etal-2024-autocap,wang2024qwen2vlenhancingvisionlanguagemodels,song2024moviechat,zhang-etal-2024-wrong,bai2025qwen3vltechnicalreport,team2025gemma,zhang-etal-2025-cchall}. As shown in Figure~\ref{fig:intro} (a), existing video LMMs typically prepend the entire video sequence to the prompt for autoregressive generation. This architecture suffers from a documented phenomenon known as \textit{\textbf{Visual Anchoring Decay}}~\cite{sun2025mitigating,fang2026seeing,du2025context}. As the reasoning chain grows, the model’s attention to initial spatiotemporal tokens progressively dilutes, triggering a drift toward linguistic priors. This leads to hallucinations and compounding errors, where reasoning steps detach from visual evidence, ultimately undermining the grounding required for video reasoning~\cite{huang2024visual,wang2024vigc,chandrasegaran2024hourvideo}.

To fill this gap, as shown in Figure~\ref{fig:intro} (b), we introduce \textbf{Latent Video Cache} (\textsc{Latent-VC}), a framework designed to sustain grounding throughout complex reasoning. Inspired by computer architecture, where prefetchers and caches bridge the gap between fast computation and slow main memory, \textsc{Latent-VC} embeds a \textit{Latent Visual Prefetcher} directly into the autoregressive process of LMMs. Rather than relying solely on repeated attention over thousands of raw video tokens (\textit{Main Memory}), our Prefetcher proactively distills and ``fetches'' anticipatory spatiotemporal semantics into learnable latent tokens (\textit{Cache}). While the raw video prefix remains in context, this persistent memory acts as a high-speed buffer that reduces direct video-token attention and counteracts \textit{Visual Anchoring Decay} by maintaining high-dimensional visual fidelity across the reasoning chain.

To prevent the \textit{Latent Visual Cache} from degenerating into uninformative placeholders, \textsc{Latent-VC} employs a two-stage training paradigm. \ding{172} First, during supervised fine-tuning, a \textit{contrastive cache alignment} objective projects the cache's hidden states back into the visual space, tightly binding cached latent tokens to key-frame embeddings and establishing a direct gradient pathway from visual semantics to the autoregressive latent space. \ding{173} Second, a vision-grounded Reinforcement Learning~(RL) stage introduces a novel \textit{latent grounding reward} to further enhance visual consistency. This reward measures key-frame coverage over completion-token hidden states, providing a trajectory-level preference signal that sustains high-fidelity visual anchoring throughout complex reasoning tasks, and effectively mitigates the issue of Visual Anchoring Decay.

Crucially, \textsc{Latent-VC} guarantees strict train-inference alignment by operating entirely on the model's native hidden states, thereby circumventing the distribution shifts that affect prior latent reasoning approaches. Instantiated on the \textit{Qwen3.5-9B}~\cite{qwen3.5}, we evaluate our framework across a suite of six benchmarks. These encompass temporal reasoning (\textit{TempCompass}~\cite{liu2024tempcompass}), multimodal comprehension (\textit{Video-MME}~\cite{fu2025video}, \textit{MVBench}~\cite{li2024mvbench}), domain-specific expertise (\textit{MMVU}~\cite{zhao2025mmvu}, \textit{Video-MMMU}~\cite{hu2025videommmu}), and spatial intelligence (\textit{VSI-Bench}~\cite{yang2024thinkingspace}). Across this diverse spectrum, \textsc{Latent-VC} consistently delivers substantial performance gains. Ultimately, these results validate our core hypothesis: proactively internalizing visual semantics via an active prefetch-and-cache memory system establishes a highly effective, foundational paradigm for long-form video reasoning.

Our main contributions are summarized as follows:

\begin{itemize}
\item [\ding{182}] \textit{\textbf{Architectural Paradigm Shift:}} We highlight the phenomenon of \textit{Visual Anchoring Decay} in video LMMs and propose \textsc{Latent-VC} as a structural solution. By introducing a \textit{Latent Visual Prefetcher and Cache}, we actively distill and anchor high-dimensional visual representations directly within the reasoning stream, bypassing lossy textual bottlenecks.
\item [\ding{183}] \textit{\textbf{Visual Prefetch Learning:}} We introduce a two-stage training framework for the latent visual cache. Stage I aligns latent-block states with key-frame embeddings. Stage II applies vision-grounded GRPO with answer, format, temporal, and latent rewards, where the latent reward measures key-frame coverage over completion-token hidden states.
\item [\ding{184}] \textit{\textbf{Stronger Performance:}} Evaluated comprehensively across six diverse benchmarks spanning temporal, spatial, and domain-specific intelligence, \textsc{Latent-VC} demonstrates consistent and significant gains. It also achieves higher accuracy with shorter responses, indicating that latent visual caching preserves visual evidence rather than longer textual chains.
\end{itemize}

To facilitate further research, all source code, trained models, and preprocessed datasets will be made fully publicly accessible at \url{https://github.com/BRZ911/Latent-VC}.

\begin{figure}[t]
	\centering
\includegraphics[width=\textwidth]{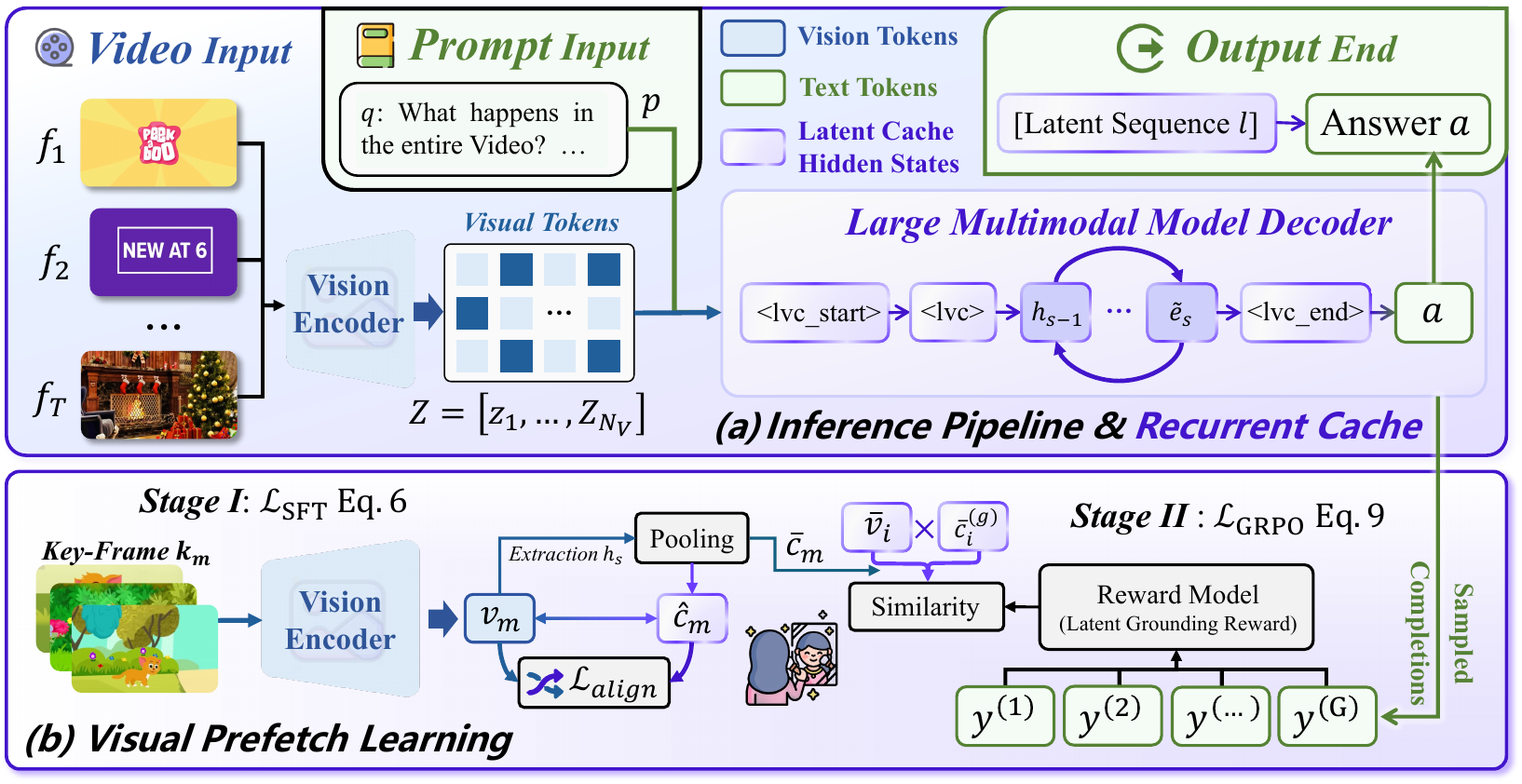}
\caption{Overall framework of Latent Video Cache. The method consists of two main components: (a) \textit{Inference Pipeline \& Recurrent Cache} and (b) \textit{Latent Prefetch Learning}.\vspace{-2mm}}
\label{fig:framework}
\end{figure}

\section{Latent Video Cache}

As shown in Figure~\ref{fig:framework}, \textsc{Latent-VC} inserts a compact latent cache into a video LMM and optimizes it with visual supervision in the hidden space. Section~\ref{sec:framework} defines the formulation and latent-cache interface. Section~\ref{sec:architecture} presents the recurrent cache mechanism. Section~\ref{sec:optimization} introduces the two-stage optimization. Section~\ref{sec:inference} specifies the inference pipeline and train-inference consistency.

\subsection{Framework Overview and Latent Cache Interface}
\label{sec:framework}

As shown in Figure~\ref{fig:framework} (a), let $V=\{f_t\}_{t=1}^{T}$ be a video with $T$ frames and let $q$ be the user question. Following the implementation, we first sample a shorter clip $\tilde V$ and feed it to the frozen visual tower plus visual merger. The resulting video-prefix tokens are:
\begin{equation}
\mathbf{Z} = \mathcal{E}_{v}(\tilde V) = [\mathbf{z}_1,\dots,\mathbf{z}_{N_v}],
\qquad \mathbf{z}_n \in \mathbb{R}^{d},
\label{eq:video_tokens}
\end{equation}
where $\mathcal{E}_{v}$ denotes the video encoder used by the backbone, $N_v$ is the number of visual tokens after visual merging, and $d$ is the decoder hidden size. In code, these vectors replace the placeholder video-token embeddings before the language model’s forward pass.

Let $\mathbf{p}=[p_1,\dots,p_{L_p}]$ be the prompt built from $(V,q)$ and let $\mathbf{a}=[a_1,\dots,a_{L_a}]$ be answer tokens. Standard video LMMs generate the answer sequence $\mathbf{a}$ directly from $[\mathbf{Z},\mathbf{p}]$. We instead insert $S$ latent-cache tokens $\boldsymbol{\ell}=[\ell_1,\dots,\ell_S]$ between the prompt and the answer, where each $\ell_s$ is the special token \texttt{<|lvc|>} and its hidden state is used as a non-verbal visual memory. These latent slots are deterministic recurrent computation states rather than answer tokens sampled from the vocabulary, so the probabilistic factorization applies only to the discrete answer tokens:
\begin{equation}
\mathbf{H}_{1:S}=\mathrm{Rollout}_{\theta}(\mathbf{Z},\mathbf{p},S),
\qquad
\pi_{\theta}(\mathbf{a}\mid V,q)
= \prod_{t=1}^{L_a} \pi_{\theta}(a_t \mid \mathbf{Z}, \mathbf{p}, \mathbf{H}_{1:S}, a_{<t}),
\label{eq:full_factorization}
\end{equation}
where $\mathbf{H}_{1:S}=\{\mathbf{h}_s\}_{s=1}^{S}$ denotes the latent-cache hidden states induced by $\boldsymbol{\ell}$, $\pi_{\theta}(\mathbf{a}\mid V,q)$ is the conditional probability of generating answer $\mathbf{a}$ given video $V$ and question~$q$, and $a_{<t}$ is the answer-token history. The latent tokens are not rationales; their states are used for visual grounding.

During supervised training, samples may include $M$ key frames $\mathcal{K}=\{k_m\}_{m=1}^{M}$. We partition the $S$ \texttt{<|lvc|>} positions into blocks and supervise each block with its key frame; Appendix~\ref{app:sft} provides details of this training-only allocation. Inference uses no key-frame annotations or block assignments.
\textsc{Latent-VC} combines components: a frozen visual prefix $\mathbf{Z}$, a recurrent latent cache implemented by the \texttt{<|lvc|>} hidden states, and a two-stage training objective. The interface uses \texttt{<|lvc\_start|>}, repeated \texttt{<|lvc|>} slots, and \texttt{<|lvc\_end|>}. Appendix~\ref{app:core} provides the execution view.

\subsection{Recurrent Latent Visual Cache}
\label{sec:architecture}

As shown in the Large Language Model Decoder module of Figure~\ref{fig:framework} (a), \textsc{Latent-VC} inserts a latent cache between the prompt and answer, unrolling recurrent hidden states before generation. Let $\mathbf{e}(x)$ be $x$'s embedding and $\mathbf{h}_t \in \mathbb{R}^{d}$ the decoder hidden state. For slot $s$, we use the hidden state as input:
\begin{equation}
\tilde{\mathbf{e}}_s =
\begin{cases}
\mathbf{e}(\ell_1), & s = 1,\\
\mathbf{h}_{s-1}, & s > 1,
\end{cases}
\qquad
\mathbf{h}_s = \mathcal{D}_{\theta}(\tilde{\mathbf{e}}_s, \mathbf{C}_{<s}),
\label{eq:latent_recurrence}
\end{equation}
where $\mathcal{D}_{\theta}$ denotes the decoder transformation and $\mathbf{C}_{<s}$ is the autoregressive context, including the visual prefix. Thus, the latent segment forms a recurrent chain within the decoder without removing raw video tokens. This maintains a continuous state, incurring modest overhead since $S \ll N_v$.

\subsection{Visual Prefetch Learning}
\label{sec:optimization}

\paragraph{Stage I: Supervised Cache Alignment.}
Stage I, as shown on the left of Figure~\ref{fig:framework} (b), aligns each latent block with key frames by using frozen visual features as targets and projected \texttt{<|lvc|>} hidden states as cache predictions. For the $m$-th matched key frame in sample $i$, we construct them as:
\begin{equation}
\mathbf{v}_{i,m} = \frac{1}{N_{i,m}} \sum_{j=1}^{N_{i,m}} \mathbf{u}_{i,m,j},
\qquad
\mathbf{c}_{i,m} = \frac{1}{|\mathcal{I}_{i,m}|} \sum_{s \in \mathcal{I}_{i,m}} \mathbf{h}_{i,s},
\qquad
\hat{\mathbf{c}}_{i,m} = P_{\phi}(\mathbf{c}_{i,m}),
\label{eq:sft_cache_target}
\end{equation}
where $\mathbf{v}_{i,m}$ is the \emph{visual target} for the $m$-th key frame, obtained by mean-pooling frozen visual patch features $\{\mathbf{u}_{i,m,j}\}_{j=1}^{N_{i,m}}$ and fixed during Stage-I training. In contrast, $\mathbf{c}_{i,m}$ is the \emph{latent-cache summary} computed by averaging decoder hidden states $\{\mathbf{h}_{i,s}:s\in\mathcal{I}_{i,m}\}$ at the \texttt{<|lvc|>} positions $\mathcal{I}_{i,m}$ assigned to that frame. Since $\mathbf{c}_{i,m}$ and $\mathbf{v}_{i,m}$ lie in different spaces, a trainable projection head $P_{\phi}$ maps $\mathbf{c}_{i,m}$ to $\hat{\mathbf{c}}_{i,m}$, the \emph{projected representation}. Thus, Eq.~\eqref{eq:sft_cache_target} constructs a matched pair $(\hat{\mathbf{c}}_{i,m},\mathbf{v}_{i,m})$, enabling Stage I to supervise the cache representation itself instead of just final answer tokens.
The central objective of Stage I is the key-frame-to-cache contrastive alignment loss:
\begin{equation}
\mathcal{L}_{\mathrm{align}}
= - \frac{1}{M^{\ast}} \sum_{m=1}^{M^{\ast}}
\log
\frac{\exp(\mathrm{sim}(\hat{\mathbf{c}}_m, \mathbf{v}_m)/\tau)}
{\sum_{n=1}^{M^{\ast}} \exp(\mathrm{sim}(\hat{\mathbf{c}}_m, \mathbf{v}_n)/\tau)},
\label{eq:align_loss}
\end{equation}
where $M^{\ast}$ denotes the number of valid matched latent--key-frame pairs in a mini-batch. For the $m$th pair, $\mathbf{v}_m$ is the frozen key-frame visual target, $\hat{\mathbf{c}}_m=P_{\phi}(\mathbf{c}_m)$ is the projected latent-cache representation, $P_{\phi}$ is the trainable projection head, $\tau$ is the contrastive temperature, and $\mathrm{sim}(\cdot,\cdot)$ denotes cosine similarity. The numerator pulls each latent cache representation toward its matched key-frame target, while the denominator contrasts it against all other key-frame targets in the mini-batch. Thus, minimizing $\mathcal{L}_{\mathrm{align}}$ prevents the cache from becoming a generic video summary.
The full supervised objective for this cache-alignment stage is then written as:
\begin{equation}
\mathcal{L}_{\mathrm{SFT}} = \mathcal{L}_{\mathrm{ce}} + \lambda_{\mathrm{lvc}} \mathcal{L}_{\mathrm{align}},
\label{eq:sft_loss}
\end{equation}
where $\mathcal{L}_{\mathrm{ce}}$ is the masked language-modeling loss over ordinary text tokens and $\lambda_{\mathrm{lvc}}$ controls the alignment strength. We freeze the vision tower and visual merger, and optimize the language backbone together with $P_{\phi}$. Appendix~\ref{app:sft} provides the cache pooling and token masking details.

\paragraph{Stage II: Vision-Grounded GRPO.}
Stage I aligns cache representations under teacher-forced supervision, but the model must also learn to use these states during free generation. Stage II, as shown on the right of Figure~\ref{fig:framework} (b), therefore starts from the Stage I checkpoint and applies GRPO~\cite{shao2024deepseekmath}. For each training prompt $(V_i,q_i)$, the trainer samples $G$ candidate completions $\{\mathbf{y}_i^{(g)}\}_{g=1}^{G}$ from the current policy. The scalar reward for completion $g$ is the weighted sum used by the implementation:
\begin{equation}
r_i^{(g)} = w_{\mathrm{acc}} r_{i,\mathrm{acc}}^{(g)}
+ w_{\mathrm{fmt}} r_{i,\mathrm{fmt}}^{(g)}
+ w_{\mathrm{tmp}} r_{i,\mathrm{tmp}}^{(g)}
+ w_{\mathrm{lat}} r_{i,\mathrm{lat}}^{(g)},
\label{eq:reward_sum}
\end{equation}
where $r_{i,\mathrm{acc}}^{(g)}$ checks whether the final answer matches the ground truth, $r_{i,\mathrm{fmt}}^{(g)}$ checks the required \texttt{<think>} and \texttt{<answer>} format, $r_{i,\mathrm{tmp}}^{(g)}$ rewards timestamps that match annotated key-frame times within a tolerance window, and $r_{i,\mathrm{lat}}^{(g)}$ measures whether valid completion-token hidden states cover the annotated key-frame targets. The latent reward is not computed from the fixed pre-answer cache rollout. It is computed from the sampled completion trajectory, so it can differ across the $G$ completions of the same prompt. The coefficients $w_{\mathrm{acc}},w_{\mathrm{fmt}},w_{\mathrm{tmp}},w_{\mathrm{lat}}$ are the reward weights.
The key GRPO normalization step is to compare completions only within the same prompt group:
\begin{equation}
A_i^{(g)}
= \frac{r_i^{(g)} - \mu_i}{\sigma_i + \epsilon},
\qquad
\mu_i = \frac{1}{G}\sum_{g=1}^{G} r_i^{(g)},
\qquad
\sigma_i = \mathrm{Std}\big(\{r_i^{(g)}\}_{g=1}^{G}\big),
\label{eq:grpo_advantage}
\end{equation}
where $A_i^{(g)}$ is the \emph{group-relative advantage} of the $g$th completion for prompt $i$, obtained by normalizing its reward $r_i^{(g)}$ against the $G$ completions sampled for the same video-question pair. The group mean $\mu_i$ serves as the prompt-specific reward baseline, $\sigma_i$ measures the reward dispersion within the group, and $\epsilon$ prevents instability. Thus, positive or negative advantages indicate completions that are better or worse than their same-prompt alternatives, reducing prompt-level difficulty bias.

GRPO then uses these group-relative advantages in a clipped token-level policy update with optional reference-model KL regularization. The core Stage-II objective is:
\begin{equation}
\mathcal{L}_{\mathrm{GRPO}}
= - \mathbb{E}_{i,g,t}
\Big[
\min\big(
\rho_{i,t}^{(g)} A_i^{(g)},
\mathrm{clip}(\rho_{i,t}^{(g)}, 1-\epsilon_{\ell}, 1+\epsilon_{h}) A_i^{(g)}
\big)
\Big]
+ \beta \, \mathbb{E}_{i,g,t}\big[D_{\mathrm{KL}}^{\mathrm{ref}}\big],
\label{eq:grpo_loss}
\end{equation}
where the expectation is taken over prompt $i$, sampled completion $g$, and valid generated answer token $t$. The ratio $\rho_{i,t}^{(g)}$ compares the current policy probability with the sampling policy probability for token $a_{i,t}^{(g)}$, and $A_i^{(g)}$ is the group-relative advantage computed from the total reward in Eq.~\eqref{eq:reward_sum}. The clipping margins $\epsilon_{\ell}$ and $\epsilon_{h}$ bound policy updates to reinforce high-reward completions without unstable probability shifts. The optional penalty $D_{\mathrm{KL}}^{\mathrm{ref}}$, weighted by $\beta$, keeps the optimized policy close to the Stage I reference. As a detached scalar, the latent reward does not backpropagate through the cosine score, but biases the policy toward visually grounded trajectories. Appendix~\ref{app:grpo} details latent-reward coverage scoring, threshold shaping, advantage normalization, and ratio/KL terms.

\subsection{Inference Pipeline and Train-Inference Consistency}
\label{sec:inference}

At inference time, the model receives only $(V,q)$ and requires no key-frame annotations: it encodes the video into $\mathbf{Z}$, emits \texttt{<|lvc\_start|>}, evolves the latent cache via Eq.~\eqref{eq:latent_recurrence}, emits \texttt{<|lvc\_end|>}, and finally decodes the answer $\mathbf{a}$. A compact execution-level summary is provided in Appendix~\ref{app:framework}.

Train-inference consistency is strict: Stage I aligns latent blocks with key visual moments, Stage II rewards key-frame coverage over completion-token hidden states produced after the same recurrent latent update, and inference uses the same update before decoding the final answer. \textsc{Latent-VC} fundamentally improves video reasoning by directly altering the model's internal computation.

\section{Experiments}

\subsection{Experimental Settings}

\textbf{Implementation Details:} We adopt Qwen3.5-9B-Base~\cite{qwen3.5} as the primary backbone model. \textsc{Latent-VC} is trained on the video-only subsets of the Open-o3-Video dataset~\cite{open-o3-video}, and detailed source statistics for both the Stage I (SFT) and Stage II (GRPO) phases are provided in Appendix~\ref{app:data}. For the reward function, we specifically set the weights as $w_{\mathrm{acc}}=2.0, w_{\mathrm{fmt}}=0.5, w_{\mathrm{tmp}}=0.5$, and $w_{\mathrm{lat}}=1.0$. The cache-alignment weight is $\lambda_{\mathrm{lvc}}=0.1$, the contrastive temperature is $\tau=0.07$, and the latent-reward threshold is $\delta=0.2$. 
For training video preprocessing, we follow Video-R1~\cite{feng2025videor1}: videos are uniformly sampled at 1 FPS, each training clip is capped at 16 frames, and visual inputs are tokenized with a patch size of $28 \times 28$. Evaluation uses frame budgets of 16, 32, and 64 frames following the benchmark protocol. The latent reasoning step count is set to 8 during inference.

\textbf{Benchmarks and Baselines:}  Following the evaluation protocol of Video-R1~\cite{feng2025videor1}, we evaluate all methods under a unified protocol on six public benchmarks: VSI-Bench~\cite{yang2024thinkingspace}, VideoMMMU~\cite{hu2025videommmu}, MMVU~\cite{zhao2025mmvu}, MVBench~\cite{li2024mvbench}, TempCompass~\cite{liu2024tempcompass}, and VideoMME~\cite{fu2025video}. We further compare against representative prior video LMMs, including LLaMA-VID~\cite{li2023llamavid}, VideoLLaMA2~\cite{cheng2024videollama2}, LongVA-7B~\cite{zhang2024longva}, VILA-1.5-8B and VILA-1.5-40B~\cite{lin2024vila}, Video-UTR-7B~\cite{yu2025videoutr}, LLaVA-OneVision-7B~\cite{li2024llavaonevision}, Kangaroo-8B~\cite{liu2024kangaroo}, and Video-R1-7B~\cite{feng2025videor1} evaluated with 16, 32, and 64 frames. In addition, we implement the Chain-of-Thought~\cite{wei2022chain} baseline (\qcot) and the SFT~\cite{ouyang2022training}+GRPO~\cite{shao2024deepseekmath} baseline (\qsft), and compare them with \textsc{Latent-VC}-9B under the same setting.

\begin{table}[t]
\centering
\footnotesize
\setlength{\tabcolsep}{3pt}
\renewcommand{\arraystretch}{1.5}
\caption{The experimental results of Acc. (\%) on LMMs. $^\dagger$ denotes results taken from prior work~\cite{feng2025videor1}, and ``-'' indicates unavailable results. $\uparrow$ indicates the performance improvement over \qcot. To ensure reliability, the results for \textsc{Latent-VC-9B} are obtained by averaging over three runs.}
\label{tab:main-results}
\resizebox{\linewidth}{!}{
\begin{tabular}{l|c|cccccc}
\specialrule{1pt}{0pt}{0pt}
\rowcolor{TableHeaderLavender}
\textbf{Model} & \textbf{Frames} & \textbf{VSI-Bench} & \textbf{VideoMMMU} & \textbf{MMVU} & \textbf{MVBench} & \textbf{TempCompass} & \textbf{VideoMME} \\
\specialrule{1pt}{0pt}{0pt}
LLaMA-VID$^\dagger$~\cite{li2023llamavid} & - & - & - & - & 41.9 & 45.6 & - \\
\rowcolor{TableStripeSoft}
VideoLLaMA2$^\dagger$~\cite{cheng2024videollama2} & - & - & - & 44.8 & 54.6 & - & 47.9 \\
LongVA-7B$^\dagger$~\cite{zhang2024longva} & - & 29.2 & 23.9 & - & - & 56.9 & 52.6 \\
\rowcolor{TableStripeSoft}
VILA-1.5-8B$^\dagger$~\cite{lin2024vila} & - & 28.9 & 20.8 & - & - & 58.8 & - \\
VILA-1.5-40B$^\dagger$~\cite{lin2024vila} & - & 31.2 & 34.0 & - & - & - & 60.1 \\
\rowcolor{TableStripeSoft}
Video-UTR-7B$^\dagger$~\cite{yu2025videoutr} & - & - & - & - & 58.8 & 59.7 & 52.6 \\
LLaVA-OneVision-7B$^\dagger$~\cite{li2024llavaonevision} & - & 32.4 & 33.8 & 49.2 & 56.7 & - & 58.2 \\
\rowcolor{TableStripeSoft}
Kangaroo-8B$^\dagger$~\cite{liu2024kangaroo} & - & - & - & - & 61.1 & 62.5 & 56.0 \\
Video-R1-7B$^\dagger$~\cite{feng2025videor1} & 16 & 34.6 & 49.8 & 64.2 & 62.7 & 72.6 & 57.4 \\
\rowcolor{TableStripeSoft}
Video-R1-7B$^\dagger$~\cite{feng2025videor1} & 32 & 35.8 & 52.3 & 63.8 & 63.9 & 73.2 & 59.3 \\
Video-R1-7B$^\dagger$~\cite{feng2025videor1} & 64 & 37.1 & 52.4 & 63.8 & 64.8 & 73.2 & 61.4 \\
\hline
\rowcolor{TableStripeSoft}
\qcot & 16 & 29.4 & 41.2 & 61.6 & 51.8 & 66.6 & 51.0 \\
\qcot & 32 & 33.4 & 42.8 & 59.8 & 52.2 & 66.5 & 50.3 \\
\rowcolor{TableStripeSoft}
\qcot & 64 & 27.9 & 44.0 & 61.6 & 52.0 & 66.5 & 48.5 \\
\hline
\rowcolor[rgb]{.910,.965,.925}
\qsft & 16 & 47.2\gaingreen{17.8} & 59.8\gaingreen{18.6} & 65.9\gaingreen{4.3} & 60.7\gaingreen{8.9} & 72.2\gaingreen{5.6} & 60.3\gaingreen{9.3} \\
\rowcolor{TableStripeGreen}
\qsft & 32 & 50.0\gaingreen{16.6} & 60.3\gaingreen{17.5} & 66.6\gaingreen{6.8} & 60.7\gaingreen{8.5} & 72.2\gaingreen{5.7} & 61.6\gaingreen{11.3} \\
\rowcolor[rgb]{.910,.965,.925}
\qsft & 64 & 52.6\gaingreen{24.7} & 59.7\gaingreen{15.7} & 67.2\gaingreen{5.6} & 60.8\gaingreen{8.8} & 72.2\gaingreen{5.7} & 64.6\gaingreen{16.1} \\
\hline
\rowcolor{TableStripePurple}
\latentcachemodel & 16 & 55.5\gainblue{26.1} & 65.2\gainblue{24.0} & 66.6\gainblue{5.0} & 62.2\gainblue{10.4} & 72.6\gainblue{6.0} & 61.9\gainblue{10.9} \\
\rowcolor[rgb]{.975,.965,.995}
\latentcachemodel & 32 & 60.4\gainblue{27.0} & 65.4\gainblue{22.6} & 67.7\gainblue{7.9} & 61.9\gainblue{9.7} & 72.6\gainblue{6.1} & 64.0\gainblue{13.7} \\
\rowcolor{TableStripePurple}
\latentcachemodel & 64 & 61.9\gainblue{34.0} & 65.3\gainblue{21.3} & 68.6\gainblue{7.0} & 62.5\gainblue{10.5} & 72.6\gainblue{6.1} & 66.1\gainblue{17.6} \\
\specialrule{1pt}{0pt}{0pt}
\end{tabular}
}
\vspace{-3mm}
\end{table}

\subsection{Main Results}

The experimental results are shown in Table~\ref{tab:main-results}.
Based on these results, we make four observations:

\textbf{Obs. 1. \textsc{Latent-VC} consistently outperforms baselines across the six diverse benchmarks.} Table~\ref{tab:main-results} shows that \textsc{Latent-VC} surpasses \qcot on all six benchmarks under various input scales of 16, 32, and 64 frames, with average gains of 13.7, 14.5, and 16.1 points, respectively. This consistent advantage indicates that the proposed latent cache improves video reasoning quality robustly across settings rather than only under specific benchmarks or frame budgets.

\textbf{Obs. 2. The gains persist even against a substantially stronger training baseline.} Relative to \qsft, \textsc{Latent-VC} improves performance on every benchmark--budget pair, with consistent average gains of 3.0, 3.4, and 3.3 points at 16, 32, and 64 frames, respectively, across all six benchmarks. These results suggest that the performance improvements are not merely a byproduct of stronger optimization alone, but instead stem directly from the recurrent latent cache itself.

\textbf{Obs. 3. \textsc{Latent-VC} is especially effective on grounding-intensive video reasoning benchmarks.} The largest improvements are consistently observed on VSI-Bench, VideoMMMU, and VideoMME, among all evaluated benchmarks, where \textsc{Latent-VC} achieves substantial gains of up to +34.0, +24.0, and +17.6 over \qcot, respectively. This pattern further indicates that maintaining a recurrent latent visual workspace is particularly beneficial for tasks that require sustained grounding over temporally distributed visual evidence across long, complex reasoning chains.

\textbf{Obs. 4. The advantage of \textsc{Latent-VC} becomes more pronounced as the visual budget increases.} In particular, the average gain over \qcot rises from 13.7 at 16 frames to 16.1 at 64 frames. On benchmarks such as VSI-Bench and VideoMME, \textsc{Latent-VC} continues to improve as more frames are provided, reaching 61.9 and 66.1 at 64 frames, whereas the \qcot baseline drops to 27.9 and 48.5. This trend suggests that the recurrent latent cache more effectively exploits additional visual evidence than conventional read-once generation.

\begin{figure}[t]
	\centering
\includegraphics[width=\textwidth]{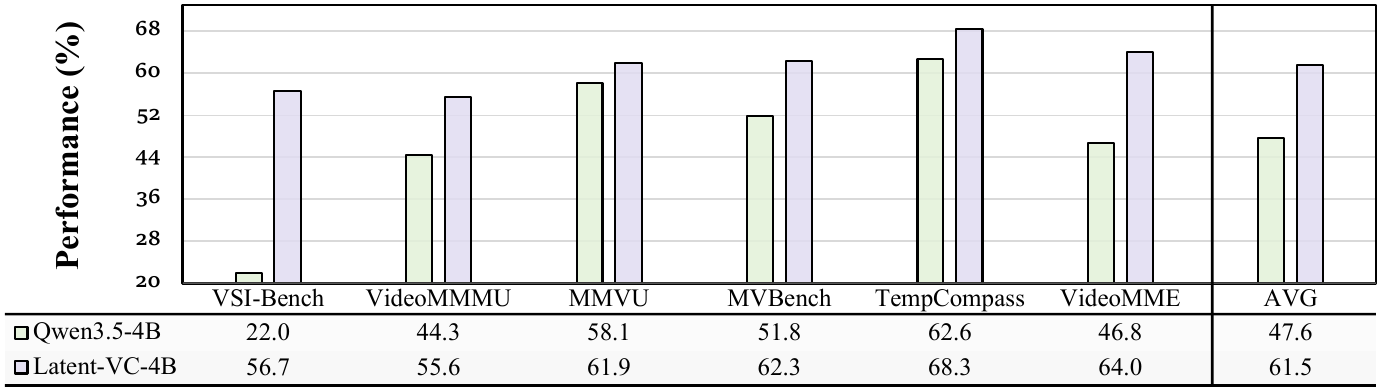}
\vspace{-3mm}
\caption{Performance comparison between Qwen3.5-4B and \textsc{Latent-VC-4B} under the 64-frames.}
\label{fig:4b-comparison}
\end{figure}

\subsection{Latent Video Cache Analysis}

To further validate the robustness of our method and better understand the source of its effectiveness, we conduct more in-depth experiments and analyses in this section.

\textbf{1. Even with the same training setup, \textsc{Latent-VC} remains effective on a smaller backbone.} To verify that the latent cache does not rely on the 9B scale, we repeat the experiment on Qwen3.5-4B with the same training settings as \textsc{Latent-VC-9B}. As shown in Figure~\ref{fig:4b-comparison}, \textsc{Latent-VC-4B} improves the average score from 47.6 to 61.5 under the 64-frame setting, with an average performance improvement of 13.9 across the six benchmarks. The largest performance improvements appear on VSI-Bench, VideoMME, and VideoMMMU, with increases of 34.7, 17.2, and 11.3, respectively. These results show that the recurrent latent cache remains effective beyond the 9B scale and is particularly beneficial for grounding-intensive video reasoning on smaller backbone.

\begin{figure}[t]
	\centering
\includegraphics[width=\textwidth]{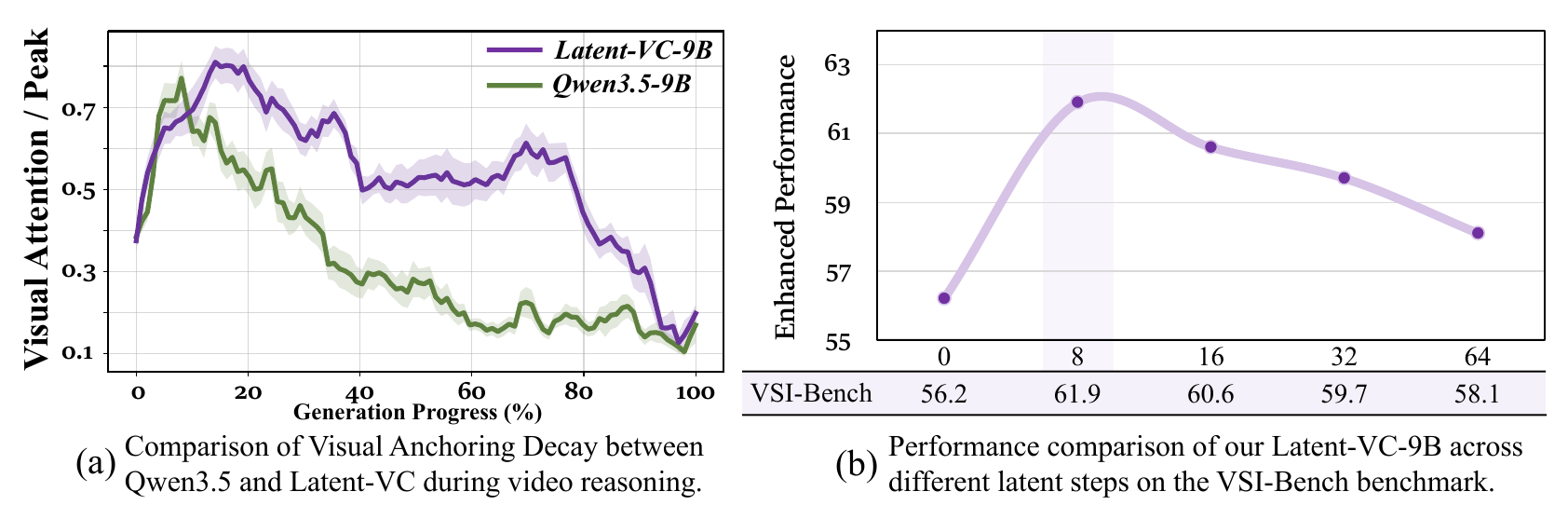}
\caption{Experiments on (a) visual anchoring dynamics during autoregressive video reasoning and (b) the effect of different latent-step configurations on VSI-Bench performance of \textsc{Latent-VC}.}
\label{fig:anchoring-analysis}
\end{figure}

\textbf{2. \textsc{Latent-VC} slows down visual anchoring decay during generation.} To verify that the gains of \textsc{Latent-VC} stem from stronger visual grounding, we compare generation-time attention mass over video tokens between \textsc{Latent-VC-9B} and Qwen3.5-9B with CoT prompting. For each token, we average attention mass across layers and heads, normalize by the peak, and aggregate it over generation progress. As shown in Figure~\ref{fig:anchoring-analysis} (a), both models start similarly, but the baseline decays much faster, whereas \textsc{Latent-VC} maintains a higher attention plateau during the middle and late stages. Head-level visual attention maps are provided in Appendix~\ref{app:head_attention}. This confirms that the recurrent latent cache mitigates \textit{Visual Anchoring Decay} throughout reasoning.

\textbf{3. A moderate number of latent steps yields the best performance.} To study how recurrent depth affects the effectiveness of the latent cache, we vary the number of latent steps while keeping the remaining settings fixed. As shown in Figure~\ref{fig:anchoring-analysis} (b), performance on VSI-Bench increases from 56.2 at 0 steps to 61.9 at 8 steps, and then gradually decreases to 60.6, 59.7, and 58.1 at 16, 32, and 64 steps, respectively. This trend suggests that a moderate recurrent depth is sufficient to refine the latent cache, whereas too many update steps may introduce redundant computation and reduce effectiveness.

\begin{table}[t]
\centering
\footnotesize
\setlength{\tabcolsep}{2pt}
\renewcommand{\arraystretch}{1.4}
\caption{Benchmark accuracy together with average response length across compared methods under the 64-frame setting. More detailed benchmark-wise results are provided in Table~\ref{tab:appendix-token-cost}.}
\label{tab:acc-len-tradeoff-64f}
\resizebox{\textwidth}{!}{
\begin{tabular}{lcccccc}
\specialrule{1pt}{0pt}{0pt}
\rowcolor{TableHeaderLavender}
\textbf{Metric} & \textbf{VSI-Bench} & \textbf{VideoMMMU} & \textbf{MMVU} & \textbf{MVBench} & \textbf{TempCompass} & \textbf{VideoMME} \\
\specialrule{1pt}{0pt}{0pt}
\rowcolor[rgb]{.975,.965,.995}
\multicolumn{1}{l}{\textbf{\textit{Accuracy}}} & \multicolumn{6}{r}{\textit{$\uparrow$ denotes the performance gain in accuracy}} \\
\hline
\qcot & 27.9 & 44.0 & 61.6 & 52.0 & 66.5 & 48.5 \\
\qsft & 52.6 & 59.7 & 67.2 & 60.8 & 72.2 & 64.6 \\
\textbf{\textsc{Latent-VC-9B}} & 61.9 & 65.3 & 68.6 & 62.5 & 72.6 & 66.1 \\
\hline
$\Delta$ vs \qcot & \textcolor{blue!65!black}{+34.0} & \textcolor{blue!65!black}{+21.3} & \textcolor{blue!65!black}{+7.0} & \textcolor{blue!65!black}{+10.5} & \textcolor{blue!65!black}{+6.1} & \textcolor{blue!65!black}{+17.6} \\
 & \textcolor{blue!65!black}{$\uparrow 121.9\%$} & \textcolor{blue!65!black}{$\uparrow 48.4\%$} & \textcolor{blue!65!black}{$\uparrow 11.4\%$} & \textcolor{blue!65!black}{$\uparrow 20.2\%$} & \textcolor{blue!65!black}{$\uparrow 9.2\%$} & \textcolor{blue!65!black}{$\uparrow 36.3\%$} \\
\specialrule{1pt}{0pt}{0pt}
\rowcolor[rgb]{.975,.965,.995}
\multicolumn{1}{l}{\textbf{\textit{Response Length}}} & \multicolumn{6}{r}{\textit{$\downarrow$ represents the reduction in generated tokens}} \\
\hline
\qcot & 1209.3 & 1552.7 & 1502.7 & 926.3 & 931.6 & 1258.3 \\
\qsft & 960.8 & 1219.9 & 829.7 & 641.6 & 594.0 & 644.4 \\
\textbf{\textsc{Latent-VC-9B}} & 463.8 & 790.4 & 564.8 & 405.9 & 424.6 & 458.2 \\
\hline
$\Delta$ vs \qcot & \textcolor{blue!65!black}{-745.5} & \textcolor{blue!65!black}{-762.3} & \textcolor{blue!65!black}{-937.9} & \textcolor{blue!65!black}{-520.4} & \textcolor{blue!65!black}{-507.0} & \textcolor{blue!65!black}{-800.1} \\
 & \textcolor{blue!65!black}{$\downarrow 61.6\%$} & \textcolor{blue!65!black}{$\downarrow 49.1\%$} & \textcolor{blue!65!black}{$\downarrow 62.4\%$} & \textcolor{blue!65!black}{$\downarrow 56.2\%$} & \textcolor{blue!65!black}{$\downarrow 54.4\%$} & \textcolor{blue!65!black}{$\downarrow 63.6\%$} \\
\specialrule{1pt}{0pt}{0pt}
\end{tabular}
}
\end{table}

\begin{figure}[t]
	\centering
\includegraphics[width=\textwidth]{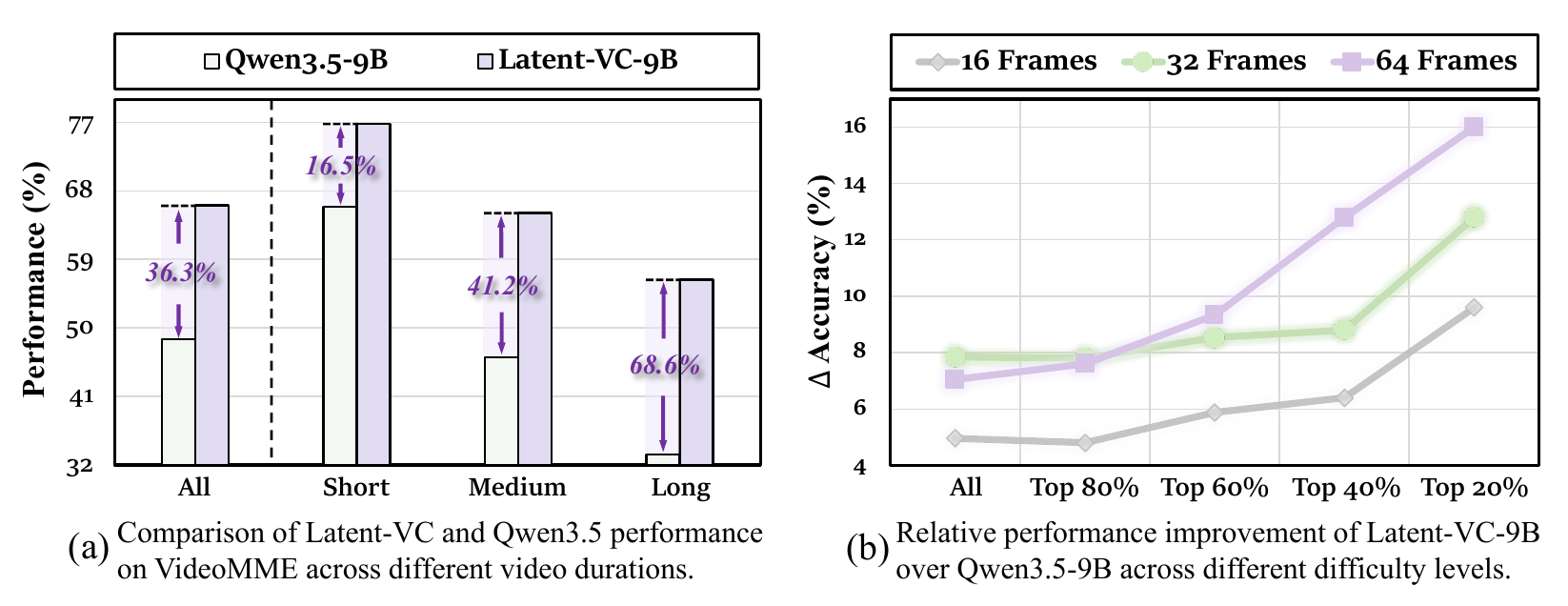}
\caption{Experiments on (a) Performance across different video durations and (b) The performance improvement of \textsc{Latent-VC-9B} over Qwen3.5-9B across different question difficulty levels. \vspace{-2mm}}
\label{fig:length-difficulty-analysis}
\end{figure}

\textbf{4. \textsc{Latent-VC} becomes more advantageous on longer videos and harder questions.} To further examine where the latent cache is most beneficial, we analyze performance by video duration on VideoMME and by question difficulty on VSI-Bench. As shown in Figure~\ref{fig:length-difficulty-analysis} (a), the improvement over Qwen3.5-9B becomes much larger on longer videos, with gains of +41.2\% and +68.6\% on the medium and long subsets, respectively. Figure~\ref{fig:length-difficulty-analysis} (b) shows a similar trend on VSI-Bench, where Top 80\%, Top 60\%, Top 40\%, and Top 20\% denote the hardest 80\%, 60\%, 40\%, and 20\% questions, respectively. The relative gains of \textsc{Latent-VC} increase on more difficult subsets, with the largest improvement observed under the 64-frame setting on the hardest split. These results suggest that even under very long token inputs, the recurrent latent cache can effectively focus on the most informative visual evidence and convert it into larger reasoning gains.

\textbf{5. \textsc{Latent-VC} improves reasoning efficiency rather than relying on longer responses.} To test whether the gains simply come from generating more text, we compare benchmark accuracy and average response length under the 64-frame setting. As shown in Table~\ref{tab:acc-len-tradeoff-64f}, \textsc{Latent-VC-9B} achieves the best accuracy on all six benchmarks while producing substantially shorter responses than \qcot, with gains of +6.1 to +34.0 points and 49.1\%--63.6\% shorter outputs. It is also both more accurate and more concise than \qsft. The same trend remains consistent across the 16-, 32-, and 64-frame settings (Appendix~\ref{app:results}), suggesting that the latent cache improves video reasoning by enhancing visual grounding rather than longer textual reasoning chains.

\begin{figure}[t]
	\centering
	\includegraphics[width=\textwidth]{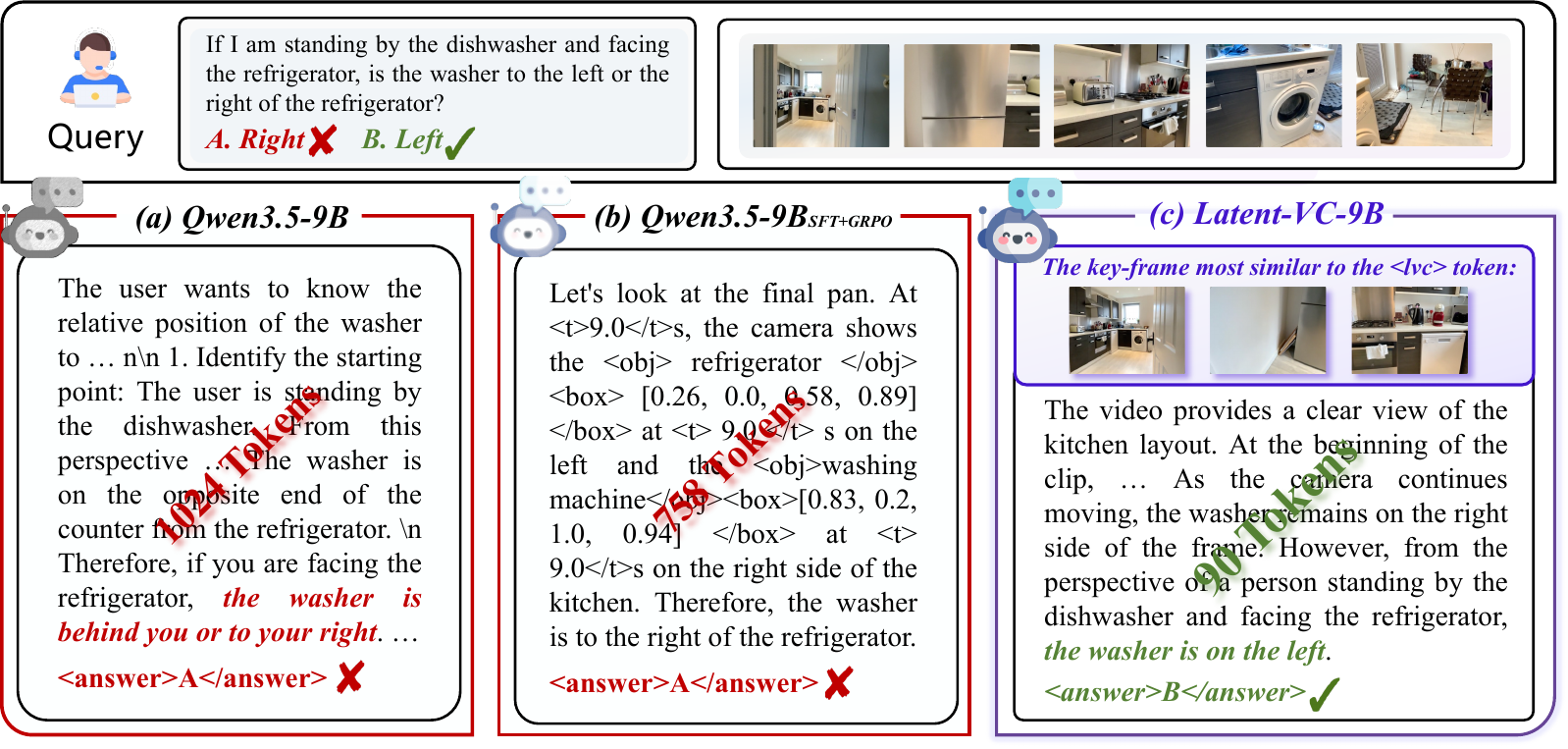}
	\vspace{-3mm}
	\caption{A case study in which Qwen3.5-9B and Qwen3.5-9B$_{\textit{SFT+GRPO}}$ both select incorrect answers, while \textsc{Latent-VC-9B} identifies the key visual evidence and predicts the correct answer. \vspace{-3mm}}
	\label{fig:case-study}
\end{figure}

\textbf{6. Qualitative analysis shows that \textsc{Latent-VC} improves visual grounding in complex reasoning.} To provide a clearer understanding of how the latent cache works in practice, we present a case study in Figure~\ref{fig:case-study}. The query asks whether the washer is on the left or right of the refrigerator from the perspective of a person standing by the dishwasher and facing the refrigerator. In Figure~\ref{fig:case-study}~(a), Qwen3.5-9B directly predicts the wrong option after a 1024-token CoT response. In Figure~\ref{fig:case-study}~(b), Qwen3.5-9B$_{\textit{SFT+GRPO}}$ also fails with a 758-token explanation containing explicit object references. In contrast, in Figure~\ref{fig:case-study}~(c), \textsc{Latent-VC-9B} identifies the key visual evidence via the latent cache and correctly infers the spatial relation using only 90 tokens. This case further highlights that the latent cache improves both the accuracy and interpretability of video reasoning.

\section{Related Work}

\noindent\textbf{Video Reasoning.} Recent progress in multimodal large language models has advanced video understanding and long-video processing~\cite{lin2024video,cheng2024videollama2,zhang2024longva,liu2024kangaroo,song2024moviechat,geminiteam2025geminifamilyhighlycapable,zhang2026thinking}. More recent works further study reasoning over video, including Video-of-Thought~\cite{fei2024video}, video-text interleaved reasoning~\cite{zhang2025vitcot}, Open-o3-Video~\cite{open-o3-video}, and Video-R1~\cite{feng2025videor1}. These studies increasingly emphasize temporal abstraction, event ordering, and evidence selection across dense visual streams. They also motivate stronger mechanisms for retaining dynamic context. However, most still follow the read-once, generate-many pipeline, making it difficult to preserve early visual evidence throughout long reasoning chains.

\noindent\textbf{Latent Reasoning.} Latent reasoning moves intermediate reasoning from text into continuous hidden states~\cite{hao2024training,xu-etal-2025-softcot,yue2026hybrid,chen2025reasoning,wang2026monet,geiping2026scaling}. This line of work suggests that continuous states can carry task-relevant computation without explicit verbalization during inference. Such compact computation is attractive for multimodal generation. COCONUT~\cite{hao2024training} and SoftCoT~\cite{xu-etal-2025-softcot} demonstrate its promise for efficient reasoning, while prior multimodal work highlights preserving visual conditioning during long CoT reasoning~\cite{sun2025mitigating}. In contrast to prior work, we study latent reasoning for video LMMs and implement it as a recurrent latent visual cache that preserves visual grounding under train-inference consistency.

\section{Conclusion}

In this work, we introduce \textsc{Latent-VC}, a latent reasoning paradigm for video understanding that maintains visual grounding through a recurrent latent visual cache. Specifically, we identify the problem of \textit{Visual Anchoring Decay}, design a latent visual prefetcher and recurrent cache within the decoder, and optimize them with supervised cache alignment and a vision-grounded RL objective. Extensive experiments show that \textsc{Latent-VC} consistently outperforms strong baselines across six benchmarks, with especially clear gains on longer and more challenging videos. These results suggest that latent-space visual caching is an effective direction for grounded long-form video reasoning.

\newpage

\bibliography{ref}
\bibliographystyle{unsrt}

\newpage

\appendix

\section{Appendix Overview}

This appendix documents implementation details omitted from the main paper for space and reproducibility. Section~\ref{app:framework} presents details of the \textsc{Latent-VC} framework, including the core execution path in Section~\ref{app:core}, the stage-one supervised cache-alignment recipe in Section~\ref{app:sft}, and the stage-two vision-grounded GRPO recipe in Section~\ref{app:grpo}. Section~\ref{app:data} details the training data and experimental settings. Section~\ref{app:results} reports additional experimental results and detailed analyses, including head-level visual attention analysis in Section~\ref{app:head_attention} and accuracy--efficiency trade-offs in Section~\ref{app:acc_efficiency}. Section~\ref{discussion} discusses broader impacts, limitations, and future directions. The algorithmic descriptions below follow the released implementation closely for faithful comparison.

\section{Latent-VC Framework Details}
\label{app:framework}

\subsection{Latent-VC Core Execution}
\label{app:core}

The core computation first converts the sampled video into a frozen visual prefix, then runs a fixed number of recurrent latent-cache steps before ordinary answer decoding. The latent slots therefore, act as intermediate visual memory states rather than readable text rationales:
\begin{equation}
\mathbf{H}_{1:S} = \mathrm{Rollout}_{\theta}(\mathbf{Z}, \mathbf{p}, S),
\qquad
\mathbf{a} \sim \pi_{\theta}(\cdot \mid \mathbf{Z}, \mathbf{p}, \mathbf{H}_{1:S}),
\label{eq:app_core_rollout}
\end{equation}
where $\mathbf{Z}$ is the frozen visual prefix, $\mathbf{p}$ is the input text prompt, $S$ is the number of \texttt{<|lvc|>} slots, $\mathbf{H}_{1:S}=\{\mathbf{h}_s\}_{s=1}^{S}$ denotes the recurrent latent-cache states, and $\mathbf{a}$ is the final decoded answer. Algorithm~\ref{alg:lvc-inference-train} provides a compact execution-level view of the proposed \textsc{Latent-VC}, covering latent-cache rollout, stage-one alignment, stage-two policy optimization, and inference-time answer generation. It should be read as the procedural counterpart to Sections~\ref{sec:framework} to~\ref{sec:inference}.

\begin{algorithm}[h]
\LinesNotNumbered
\caption{Execution Flow of \textsc{Latent-VC}}
\label{alg:lvc-inference-train}
\KwIn{Video $V$, question $q$, prompt $\mathbf{p}$, optional training annotations $\mathcal{K}=\{k_m\}_{m=1}^{M}$ and boxes}
\KwOut{Answer $\mathbf{a}$; optional training losses}
Sample clip $\tilde V$ from $V$ and encode visual tokens $\mathbf{Z} \leftarrow \mathcal{E}_v(\tilde V)$\; \algblue{Encode video prefix}
Initialize prefix state with $[\mathbf{Z}, \mathbf{p}, \texttt{<|lvc\_start|>}]$\; \algblue{Prime latent cache}
\For{$s \leftarrow 1$ \KwTo $S$}{
  \eIf{$s = 1$}{
    $\tilde{\mathbf{e}}_s \leftarrow \mathbf{e}(\ell_1)$\; \algblue{Seed first cache slot}
  }{
    $\tilde{\mathbf{e}}_s \leftarrow \mathbf{h}_{s-1}$\; \alggreen{Feed previous cache state}
  }
  $\mathbf{h}_s \leftarrow \mathcal{D}_{\theta}(\tilde{\mathbf{e}}_s, \mathbf{C}_{<s})$\; \alggreen{Advance recurrent cache}
}
Emit \texttt{<|lvc\_end|>} and decode final answer $\mathbf{a}$ autoregressively\; \alggreen{Decode grounded answer}
\If{stage = SFT}{
  Compute stage-one losses using Algorithm~\ref{alg:sft-appendix}\; \algred{Apply supervised alignment}
}
\If{stage = GRPO}{
  Compute stage-two policy update using Algorithm~\ref{alg:grpo-appendix}\; \algred{Run GRPO optimization}
}
\end{algorithm}

\subsection{Stage-I SFT Recipe}
\label{app:sft}

Stage I optimizes Eq.~\eqref{eq:sft_loss} while keeping the vision tower and visual merger frozen, so only the language backbone and latent projection head are updated. Algorithm~\ref{alg:sft-appendix} details the sequence of key-frame encoding, latent-block pooling, projection, and joint cross-entropy plus InfoNCE optimization.

For a training instance with $M_i$ annotated key frames, the implementation partitions the $S$ \texttt{<|lvc|>} positions into contiguous blocks $\{\mathcal{I}_{i,m}\}_{m=1}^{M_i}$. If $r_{i,m}$ is the number of merged visual tokens produced for key frame $k_{i,m}$, the assigned block length follows:
\begin{equation}
s_{i,m} = |\mathcal{I}_{i,m}|,
\qquad
s_{i,m} \propto r_{i,m},
\qquad
\sum_{m=1}^{M_i} s_{i,m} = S.
\label{eq:slot_budget}
\end{equation}
Where $\mathcal{I}_{i,m}$ is the designated latent block assigned to key frame $k_{i,m}$, $s_{i,m}$ is its individual length, $r_{i,m}$ is the merged visual-token count for that key frame, $M_i$ is the number of annotated key frames, and $S$ is the total number of \texttt{<|lvc|>} slots. This allocation is used only for supervised training; no such block assignment is required during the inference stage.

\begin{algorithm}[h]
\LinesNotNumbered
\caption{Stage-I Training via Supervised Cache Alignment}
\label{alg:sft-appendix}
\KwIn{Mini-batch $\{(V_i,q_i,\mathcal{K}_i,\{\mathcal{I}_{i,m}\})\}_{i=1}^{B}$}
\KwOut{Updated parameters $\theta,\phi$}
Encode each sampled video clip into visual prefix tokens $\mathbf{Z}_i \leftarrow \mathcal{E}_v(\tilde V_i)$\; \algblue{Encode video evidence}
Run the decoder with latent-cache recurrence in Eq.~\eqref{eq:latent_recurrence} and obtain hidden states $\{\mathbf{h}_{i,s}\}$\; \alggreen{Roll latent cache}
Mask cache special tokens and compute language loss $\mathcal{L}_{\mathrm{ce}}$\; \algblue{Preserve answer modeling}
\ForEach{annotated key frame $k_{i,m} \in \mathcal{K}_i$}{
  Encode key-frame tokens $\{\mathbf{u}_{i,m,j}\}_{j=1}^{N_{i,m}}$ with the frozen vision tower\; \algblue{Encode supervision frame}
  Compute visual target $\mathbf{v}_{i,m} \leftarrow \frac{1}{N_{i,m}} \sum_j \mathbf{u}_{i,m,j}$\; \algblue{Pool visual target}
  Pool the corresponding latent block $\mathbf{c}_{i,m} \leftarrow \frac{1}{|\mathcal{I}_{i,m}|} \sum_{s \in \mathcal{I}_{i,m}} \mathbf{h}_{i,s}$\; \alggreen{Pool cache evidence}
  Project latent cache state $\hat{\mathbf{c}}_{i,m} \leftarrow P_{\phi}(\mathbf{c}_{i,m})$\; \alggreen{Project into visual space}
}
Aggregate all matched pairs $\{(\hat{\mathbf{c}}_{i,m}, \mathbf{v}_{i,m})\}$ across the mini-batch\; \algblue{Assemble batch pairs}
Compute contrastive InfoNCE loss $\mathcal{L}_{\mathrm{align}}$ over matched pairs\; \algred{Align cache to key frames}
Optimize $\mathcal{L}_{\mathrm{SFT}} = \mathcal{L}_{\mathrm{ce}} + \lambda_{\mathrm{lvc}} \mathcal{L}_{\mathrm{align}}$\; \algred{Update backbone and head}
\end{algorithm}

Concretely, the comprehensive appendix algorithm presented in this work instantiates the specific visual targets, the pooled cache states, and the projected cache states as follows:
\begin{equation}
\mathbf{v}_{i,m} = \frac{1}{N_{i,m}} \sum_{j=1}^{N_{i,m}} \mathbf{u}_{i,m,j},
\qquad
\mathbf{c}_{i,m} = \frac{1}{|\mathcal{I}_{i,m}|} \sum_{s \in \mathcal{I}_{i,m}} \mathbf{h}_{i,s},
\qquad
\hat{\mathbf{c}}_{i,m} = P_{\phi}(\mathbf{c}_{i,m}),
\label{eq:app_sft_targets}
\end{equation}
where $\mathbf{v}_{i,m}$ is the frozen visual target, $N_{i,m}$ is the number of patch features $\mathbf{u}_{i,m,j}$ for key frame $k_{i,m}$, $\mathbf{c}_{i,m}$ is the average decoder state over latent block $\mathcal{I}_{i,m}$, $\mathbf{h}_{i,s}$ is the decoder hidden state at slot $s$, $P_{\phi}$ is the projection head, and $\hat{\mathbf{c}}_{i,m}$ is the projected cache state in visual feature space. 

The language-modeling loss is applied only to ordinary text tokens. Let $\mathcal{T}_{\mathrm{lvc}}$ denote the set of cache special tokens, such as \texttt{<|lvc|>}. These positions are assigned the ignore label in the implementation:
\begin{equation}
\mathcal{L}_{\mathrm{ce}}
= - \sum_{t=1}^{S+L_a}
\mathbb{1}[y_t \notin \mathcal{T}_{\mathrm{lvc}}]
\log \pi_{\theta}(y_t \mid \mathbf{Z},\mathbf{p},y_{<t}),
\label{eq:ce_loss}
\end{equation}
where $\mathcal{L}_{\mathrm{ce}}$ is the standard token-level language-modeling loss, $t$ indexes output positions, $S$ is the number of latent-cache slots, $L_a$ is the target answer length, $y_t$ is the target token at position $t$, $\mathcal{T}_{\mathrm{lvc}}$ is the set of cache special tokens ignored by the loss, $\mathbf{Z}$ is the frozen visual prefix, $\mathbf{p}$ is the input prompt, $y_{<t}$ is the preceding token context, and $\pi_{\theta}$ is the parameterized language-model policy.

These quantities are then coupled through the following mini-batch contrastive objective:
\begin{equation}
\mathcal{L}_{\mathrm{align}}^{\mathrm{batch}}
= - \frac{1}{M^{\ast}} \sum_{(i,m)}
\log
\frac{\exp(\mathrm{sim}(\hat{\mathbf{c}}_{i,m}, \mathbf{v}_{i,m})/\tau)}
{\sum_{(i',m')} \exp(\mathrm{sim}(\hat{\mathbf{c}}_{i,m}, \mathbf{v}_{i',m'})/\tau)},
\label{eq:app_sft_infonce}
\end{equation}
where $\mathcal{L}_{\mathrm{align}}^{\mathrm{batch}}$ is the mini-batch contrastive alignment loss, $M^{\ast}$ is the number of matched cache--frame pairs in the mini-batch, $(i,m)$ indexes the $m$-th annotated key frame of sample $i$, $\hat{\mathbf{c}}_{i,m}$ is the projected latent-cache state, $\mathbf{v}_{i,m}$ is its matched visual target, $(i',m')$ indexes candidate visual targets used as negatives, $\mathrm{sim}(\cdot,\cdot)$ is the similarity function, and $\tau$ is the temperature. Algorithm~\ref{alg:sft-appendix} is therefore the operational realization of Eqs.~\eqref{eq:app_sft_targets} and~\eqref{eq:app_sft_infonce}.
Two details clarify supervision: latent block $\mathcal{I}_{i,m}$ is assigned from annotations rather than learned routing, fixing which slots are supervised by key frame $k_{i,m}$. As $P_{\phi}$ maps decoder states to frozen visual space, supervision primarily shapes the decoder-side cache representation. Mini-batch negatives in Eq.~\eqref{eq:app_sft_infonce} discourage generic summaries and separate critical moments. Combined with the language loss in Eq.~\eqref{eq:ce_loss}, this yields a simple division of labor: $\mathcal{L}_{\mathrm{ce}}$ handles answer generation, while $\mathcal{L}_{\mathrm{align}}$ anchors latent states to visual evidence.

\subsection{Stage-II GRPO Recipe}
\label{app:grpo}

Stage II initializes from the Stage I checkpoint to optimize Eq.~\eqref{eq:grpo_loss}. Reward channels include answer accuracy, output format, temporal grounding, and latent grounding. Algorithm~\ref{alg:grpo-appendix} details sampling, latent-state extraction, reward construction, advantage normalization, and GRPO steps. For each prompt group, the algorithm instantiates the reward and advantage as follows:
\begin{equation}
r_i^{(g)} = \sum_{k \in \{\mathrm{acc},\mathrm{fmt},\mathrm{tmp},\mathrm{lat}\}} w_k r_{i,k}^{(g)},
\qquad
A_i^{(g)} = \frac{r_i^{(g)} - \mu_i}{\sigma_i + \epsilon},
\label{eq:app_grpo_reward_adv}
\end{equation}
where, $r_i^{(g)}$ is the total reward for completion $g$ of prompt $i$, $k$ indexes the reward channel, $w_k$ is its weight, $r_{i,k}^{(g)}$ is the channel reward, and $A_i^{(g)}$ is the normalized advantage computed with statistics $\mu_i$ and $\sigma_i$ plus stability constant $\epsilon$.
The prompt-level mean reward and reward standard deviation are
\begin{equation}
\mu_i = \frac{1}{G} \sum_{g=1}^{G} r_i^{(g)},
\qquad
\sigma_i = \mathrm{Std}\big(\{r_i^{(g)}\}_{g=1}^{G}\big),
\label{eq:app_grpo_stats}
\end{equation}
where $\mu_i$ is the mean reward over the $G$ sampled completions, and $\sigma_i$ is the standard deviation.
The latent-grounding reward measures key-frame coverage over valid completion-token hidden states. Let $\mathcal{J}_i^{(g)}$ denote valid answer-token positions after masking cache special tokens, padding, EOS, and other invalid positions. For each key frame, we search over the sampled completion trajectory:
\begin{equation}
\mathcal{J}_i^{(g)} = \{t: a_{i,t}^{(g)} \notin \mathcal{T}_{\mathrm{lvc}} \cup \{\mathrm{PAD},\mathrm{EOS}\}\},
\qquad
s_{i,m}^{(g)} = \max_{t\in \mathcal{J}_i^{(g)}} \mathrm{sim}\big(P_{\phi}(\mathbf{h}_{i,t}^{(g)}),\mathbf{v}_{i,m}\big).
\label{eq:keyframe_coverage}
\end{equation}
The completion-level grounding score then averages coverage over all annotated key frames:
\begin{equation}
s_i^{(g)} = \frac{1}{M_i}\sum_{m=1}^{M_i}s_{i,m}^{(g)}.
\label{eq:completion_pool}
\end{equation}
Where $\mathbf{h}_{i,t}^{(g)}$ is the decoder hidden state at answer-token position $t$, $P_{\phi}$ is the fixed Stage-I projection head, $\mathbf{v}_{i,m}$ is the frozen target for key frame $m$, and $M_i$ is the number of key frames. This per-key-frame coverage score avoids collapsing the trajectory into a single global summary and requires the completion to cover all annotated visual moments.

The latent-grounding reward is then computed by a thresholded cosine-similarity shaping function:
\begin{equation}
r_{i,\mathrm{lat}}^{(g)}
= \psi\big(s_i^{(g)}; \delta\big),
\label{eq:latent_reward}
\end{equation}
\begin{equation}
\psi(s;\delta)=
\begin{cases}
\frac{s-\delta}{1-\delta}, & s \ge \delta,\\
\frac{s-\delta}{\delta}, & s < \delta.
\end{cases}
\end{equation}
Thus, completions whose hidden-state trajectories cover the annotated key-frame targets more strongly than $\delta$ receive positive latent reward, while poorly grounded completions receive negative reward.

For answer token $t$ in completion $\mathbf{a}_i^{(g)}$, the policy update uses the probability ratio between the current policy and the policy that generated the sampled completion:
\begin{equation}
\rho_{i,t}^{(g)}
=
\frac{\pi_{\theta}(a_{i,t}^{(g)} \mid \mathbf{Z}_i,\mathbf{p}_i,\mathbf{H}_{i,1:S}, a_{i,<t}^{(g)})}
{\pi_{\theta_{\mathrm{old}}}(a_{i,t}^{(g)} \mid \mathbf{Z}_i,\mathbf{p}_i,\mathbf{H}_{i,1:S}, a_{i,<t}^{(g)})}.
\label{eq:importance_ratio}
\end{equation}
The implementation computes this ratio from token log probabilities and masks out padding, EOS, invalid positions, and cache special tokens in $\mathcal{T}_{\mathrm{lvc}}$; only ordinary answer tokens contribute to the GRPO objective. It also optionally adds a reference-policy penalty using the per-token estimator
\begin{equation}
D_{\mathrm{KL}}^{\mathrm{ref}}
= \exp\big(\log \pi_{\mathrm{ref}} - \log \pi_{\theta}\big)
- \big(\log \pi_{\mathrm{ref}} - \log \pi_{\theta}\big) - 1,
\end{equation}
where $\pi_{\mathrm{ref}}$ is the reference model and $\beta$ controls the penalty strength.

Finally, the token-level clipped objective minimized in Stage II is repeated below for completeness:
\begin{equation}
\mathcal{L}_{\mathrm{GRPO}}
= - \mathbb{E}_{i,g,t}
\Big[
\min\big(
\rho_{i,t}^{(g)} A_i^{(g)},
\mathrm{clip}(\rho_{i,t}^{(g)}, 1-\epsilon_{\ell}, 1+\epsilon_{h}) A_i^{(g)}
\big)
\Big]
+ \beta \, \mathbb{E}_{i,g,t}\big[D_{\mathrm{KL}}^{\mathrm{ref}}\big],
\label{eq:app_grpo_loss}
\end{equation}
where $\epsilon_{\ell}$ and $\epsilon_{h}$ are the lower and upper clipping margins. We set $\epsilon_{\ell}=\epsilon_{h}=0.2$, $\beta=0.04$, and the normalization constant $\epsilon=10^{-6}$. Taken together, these equations are translated by Algorithm~\ref{alg:grpo-appendix} into the token-level GRPO update with KL regularization over sampled responses in each group.

\begin{algorithm}[h]
\LinesNotNumbered
\caption{Stage-II Training via Vision-Grounded GRPO}
\label{alg:grpo-appendix}
\KwIn{Prompt mini-batch $\{(V_i,q_i,\mathcal{K}_i)\}_{i=1}^{B}$, stage-one policy $\pi_{\theta}$, reference policy $\pi_{\mathrm{ref}}$}
\KwOut{Updated policy parameters $\theta$}
Encode each sampled video clip into prefix tokens $\mathbf{Z}_i$\vspace{-2mm}\; \algblue{Encode visual prefix}
\ForEach{prompt $(V_i,q_i)$}{
  Sample $G$ completions $\{\mathbf{y}_i^{(g)}\}_{g=1}^{G}$ from the current policy\; \alggreen{Draw grouped rollouts}
}
\ForEach{sampled completion $\mathbf{y}_i^{(g)}$}{
  Extract valid answer-token hidden states $\{\mathbf{h}_{i,t}^{(g)}:t\in\mathcal{J}_i^{(g)}\}$ after masking cache special tokens and invalid tokens\; \algblue{Recover trajectory states}
  Compute answer-accuracy reward $r_{i,\mathrm{acc}}^{(g)}$\;
  Compute output-format reward $r_{i,\mathrm{fmt}}^{(g)}$\;
  Compute temporal-grounding reward $r_{i,\mathrm{tmp}}^{(g)}$\;
  For each key-frame target, compute its best matching projected answer-token state and average the coverage scores to obtain $r_{i,\mathrm{lat}}^{(g)}$\; \alggreen{Score visual grounding}
  Form total reward $r_i^{(g)} = \sum_k w_k r_{i,k}^{(g)}$\; \algblue{Aggregate reward channels}
}
Compute group-relative statistics $\mu_i, \sigma_i$ and normalized advantages $A_i^{(g)}$ for each prompt group\; \algblue{Normalize within prompt}
Compute answer-token ratios $\rho_{i,t}^{(g)}$ with cache special tokens masked, then apply the clipped GRPO objective with KL regularization to $\pi_{\mathrm{ref}}$\; \algred{Apply clipped GRPO}
Update $\theta$ by minimizing $\mathcal{L}_{\mathrm{GRPO}}$\; \algred{Improve policy}
\end{algorithm}

Unlike standard outcome-only rewards, the latent-grounding reward scores completion-conditioned internal trajectories rather than the fixed pre-answer cache rollout. Therefore, it is generally not constant across the $G$ completions of a prompt and is not eliminated by group-relative normalization. The reward is detached in the policy-gradient update: GRPO does not backpropagate through the cosine similarity or through $P_{\phi}$, but it effectively increases the likelihood of generating answer-token trajectories that cover all annotated key-frame targets. The KL term in Eq.~\eqref{eq:grpo_loss} further limits reward hacking by keeping the policy close to the stage-one reference model.

\section{Training Data and Experimental Settings}
\label{app:data}

\subsection{Training Data}

Our training data is curated from the released Open-o3-Video collection \cite{open-o3-video}. Because \textsc{Latent-VC} is trained for video understanding, we keep the video subsets that require temporal or temporal-spatial reasoning, and we exclude the image-only subsets, namely GQA and TreeVGR. We do not use the keyframe-only image portion of VideoEspresso. Every training sample used in both stages is associated with a full video clip. Table~\ref{tab:training_data_composition} summarizes the resulting training data composition.

\paragraph{Stage I: Supervised Fine-Tuning (SFT).}
Stage I uses a 7,047-sample corpus built entirely from the STGR split of Open-o3-Video. This split provides video clips together with annotated key frames and object bounding boxes, which we use as auxiliary supervision for latent cache alignment. The source-level distribution is as follows: PLM-RDCap (3,168), QVHighlights (1,585), ActivityNet (891), COIN (670), DiDeMo (629), and QuerYD (104). All samples are uniformly cast as temporal-spatial QA. Each training instance contains between one and five annotated key frames with spatial grounding, which supervise the latent visual cache during training but are not required at inference.

\paragraph{Stage II: Vision-Grounded GRPO.}
Stage II starts from the stage-one checkpoint and optimizes on a 32,231-sample mixture corresponding to the STGR-RL-v2 split of Open-o3-Video. This mixture combines six data components: Video-R1 multiple-choice QA (13,000), STGR temporal-spatial free-form QA (7,047), VideoEspresso full-video open-ended reasoning (5,000), TVG-R1 temporal grounding (2,904; 2,212 from QVHighlights and 692 from ActivityNet), TimeRFT temporal QA~(2,280), and Video-R1 free-form video QA (2,000). 

\begin{table}[t]
\caption{Training data composition. All retained samples come from the public Open-o3-Video~\cite{open-o3-video} collection, while raw videos are inherited from the corresponding original upstream datasets.}
\label{tab:training_data_composition}
\centering
\small
\setlength{\tabcolsep}{6pt}
\begin{tabularx}{0.98\linewidth}{@{}llXr@{}}
\toprule
Stage & Source & Task & \# Samples \\
\midrule
SFT & STGR & Temporal-spatial free-form QA & 7,047 \\
\midrule
GRPO & Video-R1 (MCQ) & General video QA (multiple-choice) & 13,000 \\
GRPO & STGR & Temporal-spatial free-form QA & 7,047 \\
GRPO & VideoEspresso & Open-ended video reasoning (full-video subset only) & 5,000 \\
GRPO & TVG-R1 & Temporal grounding & 2,904 \\
GRPO & TimeRFT & Temporal QA & 2,280 \\
GRPO & Video-R1 (Free) & General video QA (free-form) & 2,000 \\
\bottomrule
\vspace{-4mm}
\end{tabularx}
\end{table}

\subsection{Video Preprocessing and Input Budget}
Following Video-R1~\cite{feng2025videor1}, training videos are uniformly sampled at 1 FPS and capped at 16 frames per clip, while evaluation uses the benchmark-specific frame budgets of 16, 32, and 64 frames reported in Table~\ref{tab:main-results}. We use a visual patch size of $28 \times 28$. For SFT, each frame is tokenized with a budget of 128--768 visual patches, corresponding to a per-frame pixel budget of 100,352--602,112. For GRPO, we use a reduced upper bound of 512 patches per frame, corresponding to a per-frame pixel budget of 100,352--401,408, in order to control generation cost and stabilize RL training. These preprocessing settings are shared across all training sources.

\section{Experimental Results}
\label{app:results}

\begin{figure}[t]
	\centering
	\includegraphics[width=\textwidth]{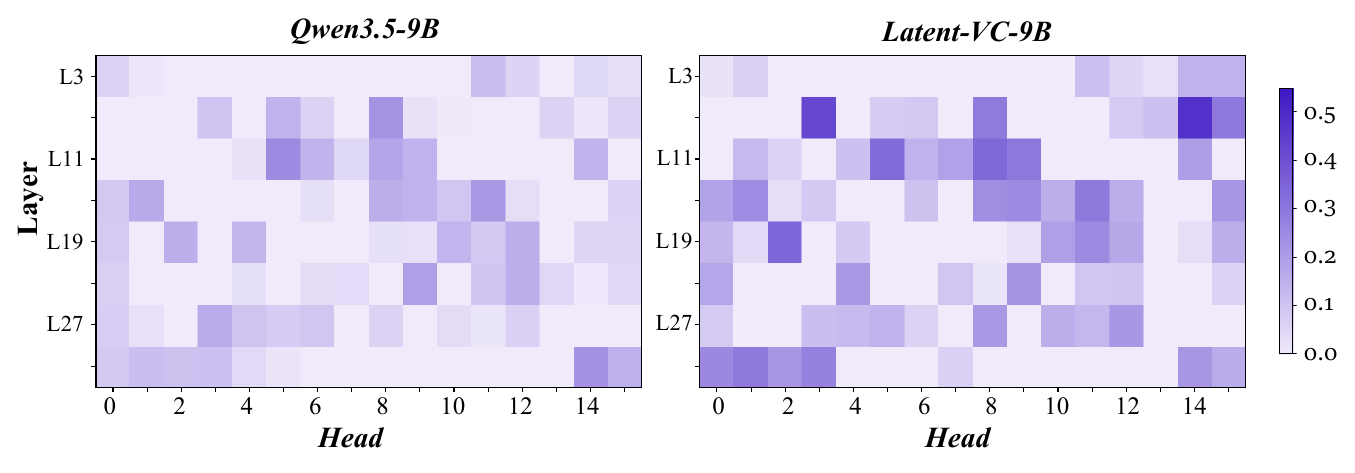}
	\vspace{-7mm}
	\caption{Head-level visual attention maps for Qwen3.5-9B and \textsc{Latent-VC-9B} during generation. Each cell reports the normalized attention mass assigned by a specific layer and head to video tokens, with darker colors indicating stronger visual anchoring.} 
	\label{fig:map}
\end{figure}

\subsection{Head-Level Visual Attention Analysis}
\label{app:head_attention}

Using token-wise attention aggregation, Figure~\ref{fig:map} provides a head-level view of visual anchoring during generation. We compare Qwen3.5-9B and \textsc{Latent-VC-9B} on the same video--question input and measure, for each generated token, how much attention each full-attention layer and head assigns to video tokens. The resulting layer--head heatmaps show that \textsc{Latent-VC-9B} produces stronger and more distributed video-token attention than the base model, especially in middle and late layers. This supports the main analysis that the recurrent latent cache mitigates \textit{Visual Anchoring Decay} by keeping visual evidence active inside the decoder during long-form reasoning.

\subsection{Accuracy--Efficiency Trade-off Across Frame Budgets}
\label{app:acc_efficiency}

\noindent\textbf{Accuracy--efficiency trade-off across frame budgets.} Appendix Table~\ref{tab:appendix-token-cost} extends the 64-frame comparison in Table~\ref{tab:acc-len-tradeoff-64f} to the 16-, 32-, and 64-frame settings and reveals three patterns. First, across all three frame budgets, \textsc{Latent-VC-9B} consistently maintains substantially shorter responses than \qcot while preserving equal or better accuracy on nearly all benchmarks, showing that gains are not purchased by longer verbal reasoning. Second, this advantage is not tied to a particular visual budget: compared with \qsft, \textsc{Latent-VC-9B} also remains shorter on every benchmark and is more accurate on nearly all benchmark--budget pairs, indicating that the recurrent latent cache improves reasoning quality rather than merely changing the training recipe. Third, the response length of \textsc{Latent-VC-9B} stays remarkably stable as the number of frames increases. For example, its outputs remain around 406 tokens on MVBench, 425 tokens on TempCompass, and 456--458 tokens on VideoMME from 16 to 64 frames, even though the model continues to improve or remain competitive in accuracy. This stability suggests that additional visual evidence is absorbed into the latent cache as compact internal computation, rather than being translated into increasingly verbose textual chains. Overall, Table~\ref{tab:appendix-token-cost} shows that \textsc{Latent-VC} achieves a strictly better accuracy--efficiency frontier across frame budgets by improving internal visual grounding.

\section{Discussion}
\label{discussion}
\noindent\textbf{Broader impacts.} \textsc{Latent-VC} provides a new perspective on grounded video reasoning by showing that visual evidence can be preserved as compact latent memories throughout autoregressive generation. This direction may benefit long-video understanding systems that require reliable temporal and spatial grounding, such as assistive agents, robotics, autonomous driving, and scientific video analysis. More broadly, by improving the accuracy--efficiency trade-off, our framework may support more concise, better grounded, and more auditable multimodal reasoning, and motivate further studies on latent visual memory and visual grounding evaluation. However, improved long-video reasoning systems may also pose risks in sensitive settings, including privacy-invasive monitoring, large-scale surveillance, and harmful decisions caused by failures in safety-critical applications such as robotics or autonomous driving, without careful validation and oversight.

\noindent\textbf{Limitations \& Future.} This work focuses on validating the effectiveness of a recurrent latent visual cache for video understanding, and several limitations remain. First, our experiments are mainly conducted on the Qwen3.5 model family; future work should examine how the approach generalizes to other backbone models, different architectures, and longer-context video LMMs. Second, the current latent grounding reward relies on available temporal-spatial supervision, which may limit its applicability to domains where such annotations are scarce or noisy. Third, while our results show that shorter responses can still achieve stronger accuracy, the interpretability of the latent cache itself remains an open problem. Future research should develop diagnostic tools for visual cache states, study robustness under noisy or streaming videos, and extend latent caching to broader scenarios.

\clearpage
\begin{table}[p]
\centering
\footnotesize
\setlength{\tabcolsep}{3pt}
\renewcommand{\arraystretch}{1.4}
\caption{Benchmark accuracy together with average response length across compared methods under the 16, 32, and 64 frame settings. $\uparrow$ and $\downarrow$ indicate the relative change over \qcot.}
\label{tab:appendix-token-cost}
\resizebox{\textwidth}{!}{%
\begin{tabular}{lcccccc}
\specialrule{1pt}{0pt}{0pt}
\rowcolor{TableHeaderLavender}
\textbf{Metric} & \textbf{VSI-Bench} & \textbf{VideoMMMU} & \textbf{MMVU} & \textbf{MVBench} & \textbf{TempCompass} & \textbf{VideoMME} \\
\hline
\rowcolor{TableHeaderLavender}
\multicolumn{7}{c}{\textbf{\textit{16 Frames}}} \\
\specialrule{1pt}{0pt}{0pt}
\rowcolor[rgb]{.975,.965,.995}
\multicolumn{7}{l}{\textbf{\textit{Accuracy}}} \\
\hline
\qcot & 29.4 & 41.2 & 61.6 & 51.8 & 66.6 & 51.0 \\
\qsft & 47.2 & 59.8 & 65.9 & 60.7 & 72.2 & 60.3 \\
\textbf{\textsc{Latent-VC-9B}} & 55.5 & 65.2 & 66.6 & 62.2 & 72.6 & 61.9 \\
\hline
$\Delta$ vs \qcot & \textcolor{blue!65!black}{+26.1} & \textcolor{blue!65!black}{+24.0} & \textcolor{blue!65!black}{+5.0} & \textcolor{blue!65!black}{+10.4} & \textcolor{blue!65!black}{+6.0} & \textcolor{blue!65!black}{+10.9} \\
 & \textcolor{blue!65!black}{$\uparrow 88.8\%$} & \textcolor{blue!65!black}{$\uparrow 58.3\%$} & \textcolor{blue!65!black}{$\uparrow 8.1\%$} & \textcolor{blue!65!black}{$\uparrow 20.1\%$} & \textcolor{blue!65!black}{$\uparrow 9.0\%$} & \textcolor{blue!65!black}{$\uparrow 21.4\%$} \\
\specialrule{0.8pt}{0pt}{0pt}
\rowcolor[rgb]{.975,.965,.995}
\multicolumn{7}{l}{\textbf{\textit{Response Length}}} \\
\hline
\qcot & 1281.5 & 1512.9 & 1423.7 & 935.2 & 928.7 & 1104.7 \\
\qsft & 1033.7 & 1349.5 & 857.0 & 651.8 & 594.4 & 657.9 \\
\textbf{\textsc{Latent-VC-9B}} & 462.7 & 823.5 & 564.3 & 406.2 & 425.0 & 457.8 \\
\hline
$\Delta$ vs \qcot & \textcolor{blue!65!black}{-818.8} & \textcolor{blue!65!black}{-689.4} & \textcolor{blue!65!black}{-859.4} & \textcolor{blue!65!black}{-529.0} & \textcolor{blue!65!black}{-503.7} & \textcolor{blue!65!black}{-646.9} \\
 & \textcolor{blue!65!black}{$\downarrow 63.9\%$} & \textcolor{blue!65!black}{$\downarrow 45.6\%$} & \textcolor{blue!65!black}{$\downarrow 60.4\%$} & \textcolor{blue!65!black}{$\downarrow 56.6\%$} & \textcolor{blue!65!black}{$\downarrow 54.2\%$} & \textcolor{blue!65!black}{$\downarrow 58.6\%$} \\
\specialrule{0.8pt}{0pt}{0pt}
\rowcolor{TableHeaderLavender}
\multicolumn{7}{c}{\textbf{\textit{32 Frames}}} \\
\specialrule{0.8pt}{0pt}{0pt}
\rowcolor[rgb]{.975,.965,.995}
\multicolumn{7}{l}{\textbf{\textit{Accuracy}}} \\
\hline
\qcot & 33.4 & 42.8 & 59.8 & 52.2 & 66.5 & 50.3 \\
\qsft & 50.0 & 60.3 & 66.6 & 60.7 & 72.2 & 61.6 \\
\textbf{\textsc{Latent-VC-9B}} & 60.4 & 65.4 & 67.7 & 61.9 & 72.6 & 64.0 \\
\hline
$\Delta$ vs \qcot & \textcolor{blue!65!black}{+27.0} & \textcolor{blue!65!black}{+22.6} & \textcolor{blue!65!black}{+7.9} & \textcolor{blue!65!black}{+9.7} & \textcolor{blue!65!black}{+6.1} & \textcolor{blue!65!black}{+13.7} \\
 & \textcolor{blue!65!black}{$\uparrow 80.8\%$} & \textcolor{blue!65!black}{$\uparrow 52.8\%$} & \textcolor{blue!65!black}{$\uparrow 13.2\%$} & \textcolor{blue!65!black}{$\uparrow 18.6\%$} & \textcolor{blue!65!black}{$\uparrow 9.2\%$} & \textcolor{blue!65!black}{$\uparrow 27.2\%$} \\
\specialrule{0.8pt}{0pt}{0pt}
\rowcolor[rgb]{.975,.965,.995}
\multicolumn{7}{l}{\textbf{\textit{Response Length}}} \\
\hline
\qcot & 1227.2 & 1550.6 & 1486.6 & 940.7 & 929.4 & 1220.0 \\
\qsft & 1019.6 & 1268.1 & 811.5 & 641.5 & 594.1 & 668.1 \\
\textbf{\textsc{Latent-VC-9B}} & 466.6 & 801.8 & 562.4 & 405.5 & 424.8 & 456.3 \\
\hline
$\Delta$ vs \qcot & \textcolor{blue!65!black}{-760.6} & \textcolor{blue!65!black}{-748.8} & \textcolor{blue!65!black}{-924.2} & \textcolor{blue!65!black}{-535.2} & \textcolor{blue!65!black}{-504.6} & \textcolor{blue!65!black}{-763.7} \\
 & \textcolor{blue!65!black}{$\downarrow 62.0\%$} & \textcolor{blue!65!black}{$\downarrow 48.3\%$} & \textcolor{blue!65!black}{$\downarrow 62.2\%$} & \textcolor{blue!65!black}{$\downarrow 56.9\%$} & \textcolor{blue!65!black}{$\downarrow 54.3\%$} & \textcolor{blue!65!black}{$\downarrow 62.6\%$} \\
\specialrule{0.8pt}{0pt}{0pt}
\rowcolor{TableHeaderLavender}
\multicolumn{7}{c}{\textbf{\textit{64 Frames}}} \\
\specialrule{0.8pt}{0pt}{0pt}
\rowcolor[rgb]{.975,.965,.995}
\multicolumn{7}{l}{\textbf{\textit{Accuracy}}} \\
\hline
\qcot & 27.9 & 44.0 & 61.6 & 52.0 & 66.5 & 48.5 \\
\qsft & 52.6 & 59.7 & 67.2 & 60.8 & 72.2 & 64.6 \\
\textbf{\textsc{Latent-VC-9B}} & 61.9 & 65.3 & 68.6 & 62.5 & 72.6 & 66.1 \\
\hline
$\Delta$ vs \qcot & \textcolor{blue!65!black}{+34.0} & \textcolor{blue!65!black}{+21.3} & \textcolor{blue!65!black}{+7.0} & \textcolor{blue!65!black}{+10.5} & \textcolor{blue!65!black}{+6.1} & \textcolor{blue!65!black}{+17.6} \\
 & \textcolor{blue!65!black}{$\uparrow 121.9\%$} & \textcolor{blue!65!black}{$\uparrow 48.4\%$} & \textcolor{blue!65!black}{$\uparrow 11.4\%$} & \textcolor{blue!65!black}{$\uparrow 20.2\%$} & \textcolor{blue!65!black}{$\uparrow 9.2\%$} & \textcolor{blue!65!black}{$\uparrow 36.3\%$} \\
\specialrule{0.8pt}{0pt}{0pt}
\rowcolor[rgb]{.975,.965,.995}
\multicolumn{7}{l}{\textbf{\textit{Response Length}}} \\
\hline
\qcot & 1209.3 & 1552.7 & 1502.7 & 926.3 & 931.6 & 1258.3 \\
\qsft & 960.8 & 1219.9 & 829.7 & 641.6 & 594.0 & 644.4 \\
\textbf{\textsc{Latent-VC-9B}} & 463.8 & 790.4 & 564.8 & 405.9 & 424.6 & 458.2 \\
\hline
$\Delta$ vs \qcot & \textcolor{blue!65!black}{-745.5} & \textcolor{blue!65!black}{-762.3} & \textcolor{blue!65!black}{-937.9} & \textcolor{blue!65!black}{-520.4} & \textcolor{blue!65!black}{-507.0} & \textcolor{blue!65!black}{-800.1} \\
 & \textcolor{blue!65!black}{$\downarrow 61.6\%$} & \textcolor{blue!65!black}{$\downarrow 49.1\%$} & \textcolor{blue!65!black}{$\downarrow 62.4\%$} & \textcolor{blue!65!black}{$\downarrow 56.2\%$} & \textcolor{blue!65!black}{$\downarrow 54.4\%$} & \textcolor{blue!65!black}{$\downarrow 63.6\%$} \\
\specialrule{1pt}{0pt}{0pt}
\end{tabular}%
}
\end{table}

\end{document}